\documentclass{article} 
\usepackage{iclr2027_conference,times}

\usepackage{amsmath,amsfonts,bm}

\def\eqref#1{equation~\ref{#1}}

\def\1{\bm{1}}

\DeclareMathAlphabet{\mathsfit}{\encodingdefault}{\sfdefault}{m}{sl}
\SetMathAlphabet{\mathsfit}{bold}{\encodingdefault}{\sfdefault}{bx}{n}

\usepackage{hyperref}
\usepackage{url}

\usepackage{microtype}
\usepackage{graphicx}
\usepackage{subcaption}
\usepackage{booktabs} 

\usepackage{algorithm}
\usepackage[noEnd=true, indLines=true, rightComments=false, beginComment=\hspace{0.5em}$\triangleright$~]{algpseudocodex}
\usepackage{float} 
\usepackage[table]{xcolor} 
\usepackage{siunitx} 
\usepackage{hyperref}

\usepackage{amsmath}
\usepackage{amssymb}
\usepackage{mathtools}
\usepackage{amsthm}
\usepackage{url}
\usepackage{listings}
\usepackage{array}
\usepackage{xspace}
\usepackage{caption}
\usepackage{multirow}
\usepackage{wrapfig}
\usepackage[most]{tcolorbox}
\usepackage{enumitem}
\usepackage{makecell}
\usepackage{xcolor}
\usepackage{placeins}

\hypersetup{
  colorlinks=true,
  citecolor=blue!60!black,
  linkcolor=black,
  urlcolor=blue!60!black
}
\usepackage{float}

\title{Learning to Simulate Individuals from\\ Macro Social Signals}
\newcommand{\xhdr}[1]{{\noindent\bfseries #1}.}
\newcommand{\modelname}{\text{macro2mind}\xspace}

\author{%
  Yining Zhao\textsuperscript{1}\thanks{Equal Contribution. Email: yiningz9@illinois.edu}  \quad  
  Bushi Liu\textsuperscript{1}\footnotemark[1]  \quad 
  Haofei Yu\textsuperscript{1}\footnotemark[1] \quad 
  Zhengyang Qi\textsuperscript{2} \quad \\
  \textbf{Shanyong Wang\textsuperscript{1} \quad
  Chuyue Li\textsuperscript{1} \quad
  Yuxiang Liu\textsuperscript{2} \quad
  Jiaxuan You\textsuperscript{1} \quad} \ \\
  \textsuperscript{1}University of Illinois Urbana-Champaign  \textsuperscript{2}Independent Researcher\\
}

\usepackage{microtype}

\usepackage{inconsolata}
\usepackage{amsmath} 

\usepackage{graphicx}
\usepackage{graphicx}
\usepackage{hyperref}
\usepackage{url}
\usepackage[utf8]{inputenc} 
\usepackage[T1]{fontenc}    
\usepackage{hyperref}       
\usepackage{url}            
\usepackage{booktabs}       
\usepackage{amsfonts}       
\usepackage{algorithm}     
\usepackage{nicefrac}       
\usepackage{microtype}      
\usepackage{xcolor}         
\usepackage{xspace}
\usepackage{tcolorbox}
\usepackage{graphicx}
\usepackage{hyperref}
 \usepackage{amsmath}
\usepackage{amsthm} 
\usepackage{bbm}
\usepackage{makecell}
\usepackage[export]{adjustbox}
\definecolor{rowgray}{gray}{0.9}   
\definecolor{rowblue}{RGB}{220,233,251}

\iclrfinalcopy 
\begin{document}

\maketitle

\begin{abstract}
Large language models are increasingly used to simulate how individuals respond to new situations, yet the behavioral reasoning behind these responses is either inherited from pretraining or learned from individual-level annotations, which offer limited behavioral diversity and little supervision of the reasoning itself. We propose to learn behavioral reasoning from prediction markets, whose price trajectories record how populations respond to real-world events at scale. We introduce \modelname, which trains a language model with GRPO using market signals. A social behavioral decomposition makes behavioral reasoning an explicit step of forecasting: the model infers representative groups of market participants, predicts how each interprets the news and updates its beliefs, reasons about their interactions, and aggregates these responses into a price. A hindsight-regret curriculum with difficulty-aware sampling focuses training on transitions where hindsight-identified groups substantially improve the forecast while prioritizing examples that remain learnable for the current policy. The learned reasoning applies to user simulation without further training. On SWM-Bench, \modelname achieves state-of-the-art directional accuracy and correlation on Polymarket. Trained on market data, it transfers zero-shot to four user-simulation benchmarks (Humanual, OvertonBench, PRISM, and CAD) and has competitive performance among zero-shot methods. Used as a data generator, \modelname also raises a downstream simulator's accuracy on unseen users by 15.5 points, outperforming data generated by its backbone by 13.2 points.
\end{abstract}

 \section{Introduction}

Large language models are increasingly used to simulate individual behavior~\citep{park2023generative, ICLR2024_b3075b88, aher2023using, ni2026survey}, including how people respond to information~\citep{argyle2023out, santurkar2023whose}, make decisions~\citep{kitadai2025can,horton2023large}, and interact with others~\citep{yu2025sotopia}. Such simulators support personalized agents~\citep{jiang2024personallm,ji2026momentum}, social simulation~\citep{yang2024oasis,gao2023s3}, and the evaluation of interactive systems~\citep{wang2025user, yao2024tau}. Achieving high-fidelity simulation of a particular individual imposes a dual requirement. It demands not only static, descriptive information about that person's background but also a dynamic capability to predict how their internal beliefs and subsequent actions evolve. We refer to this capability as \emph{behavioral reasoning}; it is closely related to theory of mind, the ability to attribute beliefs and intentions to others~\citep{premack1978does}.

To improve response fidelity, training-based simulators directly fit models to observed user behaviors~\citep{naous2026flipping, zhou2026odyssim}; some frameworks further align latent user states with these observed outputs~\citep{wu2026humanlm}. However, while rich profiles and histories successfully specify \emph{whom} to simulate, the core mechanics of \emph{how} they reason are either passively inherited from pre-training priors or explicitly fitted to individual-level annotations. Consequently, the supervision for dynamic behavioral reasoning remains strictly confined to micro-level data.

\xhdr{Challenges of user simulation} Generalizing to unseen users and situations is the central challenge for current simulators, fundamentally bottlenecked by two data limitations. (\textbf{1}) \textbf{Limited behavioral diversity.} Capturing how people dynamically update beliefs and adapt to peers requires temporally dense, socially situated data. However, such data is exceedingly scarce: traditional surveys yield only static snapshots, massive micro-level collection is costly and privacy-sensitive~\citep{gabriel2024ethics, durmus2023towards}, and synthetic personas~\citep{ge2024scaling} lack grounding in real behavior. (\textbf{2}) \textbf{Costly reasoning supervision.} Supervising \emph{how} people respond requires internal cognitive traces that are prohibitively expensive to collect. Relying on affordable alternatives introduces critical flaws: simplistic tasks encourage models to exploit superficial heuristics, while LLM-generated rationales often inject spurious reasoning disconnected from genuine cognition. Both prevent the acquisition of robust behavioral reasoning needed for generalization.

\begin{figure}[t!]
    \vspace{-4mm}
    \centering
    \includegraphics[width=\linewidth]{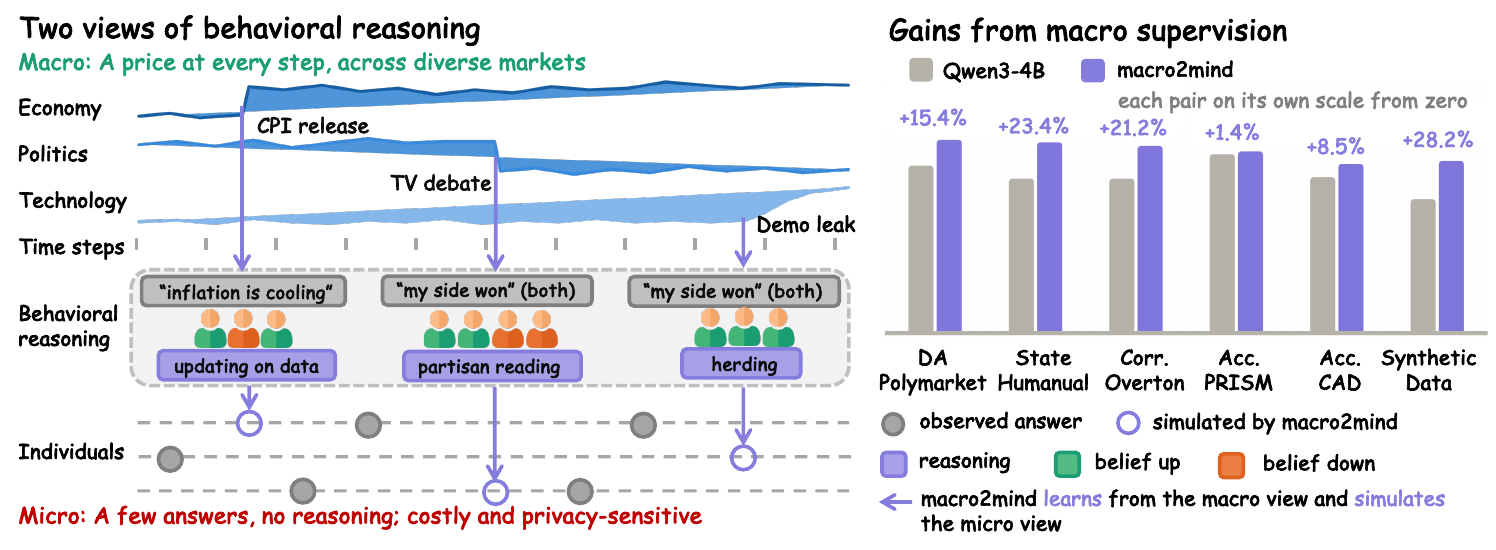}

\caption{\textbf{Learning to simulate individuals from macro signals.}
\textbf{Left:} Each prediction-market price aggregates the \emph{behavioral reasoning} of its participants: how people interpret an event given their prior beliefs and goals, update those beliefs, and respond, including to one another. Prices record such reasoning at every move, across diverse types of markets. Individual-level data offers only a few answers per person and no record of the reasoning behind them. \modelname learns behavioral reasoning from market prices and applies it zero-shot to simulate individuals.
\textbf{Right:} Relative gains of \modelname over its Qwen3-4B backbone on market forecasting and zero-shot user simulation. For synthetic data, students are trained on data generated by each model.}
    \label{fig:intro_figure}
    \vspace{-5mm}
\end{figure}

\xhdr{Market prices as aggregated individual behavior}
While individual behavioral data is scarce, prediction markets continuously record the underlying reasoning at scale~\citep{wolfers2004prediction}. Economists recognize prices not as abstract labels, but as aggregations of dispersed beliefs~\citep{hayek2013use, manski2006interpreting}. Consequently, each price movement is the macroscopic trace of a micro-level process: heterogeneous individuals interpret news through personal priors~\citep{kandel1995differential}, adapt to peers~\citep{bikhchandani1992theory}, and aggregate into a new price---mirroring the macro--micro--macro chain of social theory~\citep{coleman1990foundations}. For example, when inflation unexpectedly drops, rate-cut probabilities surge because trading desks instantly adjust, analysts react to the desks, and contrarians bet against the herd. Explaining this shift requires the exact behavioral reasoning necessary for user simulation, simply executed in reverse: inferring latent constituent behaviors from aggregate outcomes. Prediction markets thus solve both data bottlenecks (Figure~\ref{fig:intro_figure}). (\textbf{1}) \textbf{Diverse and abundant dynamics.} Markets naturally capture varied reasoning patterns from macroeconomic updates to partisan interpretations through continuous, high-stakes decisions, bypassing costly micro-level data collection. (\textbf{2}) \textbf{Outcome-grounded supervision.} The realized price provides objective outcome-level supervision for whether the aggregate implication of the inferred responses matches the observed market movement. This raises our central question: \textit{can learning to predict aggregate outcomes teach behavioral reasoning that transfers to simulating individuals?}

\xhdr{Our approach}
Harnessing this abundance is not a free lunch: evaluating only a final numerical forecast~\citep{gu2025verbalts} introduces two learning challenges. First, naive optimization risks rewarding shortcut learning over genuine behavioral reasoning (\textbf{C1}). Second, market transitions lack explicit labels to distinguish learnable social dynamics from routine noise (\textbf{C2}). We introduce \modelname, which tackles both challenges through two complementary mechanisms trained via GRPO on macroscopic rewards. First, a \textbf{social behavioral decomposition} explicitly structures the reasoning trajectory (\textbf{C1}). The model must infer representative groups, predict their belief updates and interactions, and aggregate these micro-responses into a price. Second, a \textbf{hindsight-regret curriculum with difficulty-aware sampling} performs dynamic data selection (\textbf{C2}). Given the realized outcome in hindsight, the policy names the driving groups; regret is measured by how much this hint improves its forecast. Hindsight regret identifies behaviorally informative transitions, while the difficulty-aware term increases the sampling weight of examples with useful within-group variation. Ultimately, the decomposition dictates \emph{how} the model reasons, while the curriculum dictates \emph{where} that learning is most informative.

\xhdr{Main findings} We train \modelname on prediction-market transitions from SWM-Bench and evaluate both its macro forecasting accuracy and zero-shot transfer to four micro-level benchmarks (Humanual, OvertonBench, PRISM, and CAD). Our experiments answer the central question affirmatively through three key findings:
(\textbf{1}) \textit{Superior market forecasting.} \modelname achieves state-of-the-art directional accuracy and correlations on Polymarket and highly competitive performance on Kalshi. 
(\textbf{2}) \textit{Robust zero-shot user simulation.} Despite zero exposure to individual-level annotations, \modelname consistently achieves competitive performance among zero-shot methods across the downstream benchmarks. Remarkably, it even outperforms several supervised simulators trained explicitly on the target domains.
(\textbf{3}) \textit{Scalable synthetic supervision.} Behavioral trajectories generated by our zero-shot model serve as high-quality training data. Fine-tuning downstream user simulators on these synthesized responses boosts their accuracy by 15.5 absolute points, proving \modelname to be a stronger teacher compared to its standard backbone.

\section{Related Works}
\label{sec:related_work}

\xhdr{LLM-based User Simulation} 
LLMs have been used as stand-ins for human respondents in surveys and behavioral experiments~\citep{argyle2023out,aher2023using,horton2023large}, although their default opinions often diverge from those of specific populations~\citep{santurkar2023whose,durmus2023towards}. To simulate particular individuals, existing methods condition on profiles, interviews, or interaction histories through prompting~\citep{park2023generative,park2024generative,ryan2025synthesizeme}, steer activations toward a target persona~\citep{feng2026persona}, or fine-tune on behavioral traces~\citep{naous2026flipping,wu2026humanlm,zhou2026odyssim}; reward models of human preferences, including personalized ones, learn from pairwise feedback~\citep{cai2026one, wang2026meet}. Synthetic personas~\citep{ge2024scaling} and digital-twin datasets~\citep{Yang2025TwinMarketASA} broaden coverage, but the former lack grounding in real behavior and the latter remain costly to collect. All of these approaches learn how people respond from individual-level data. We instead learn behavioral reasoning from aggregate outcomes and use user-specific data only at inference.

\xhdr{Aggregate Social Signals and Social Simulation}
Forecasting systems treat aggregate signals such as prediction-market prices as the end goal~\citep{wolfers2004prediction,halawi2024approaching,karger2025forecastbench,chen2025decisionflow}. Social simulation runs bottom-up, generating collective dynamics from rule-based~\citep{schelling1971dynamic,brock1998heterogeneous} or LLM-based agents~\citep{gao2023s3,yang2024oasis,zhang2025socioverse}, with aggregate data used only to calibrate or validate the agents~\citep{fagiolo2007critical}. Ecological inference~\citep{king1999solution} and learning from label proportions~\citep{quadrianto2008estimating} do infer individuals from aggregates, but within a fixed population and label space. We instead use aggregate outcomes to train a reasoning process that transfers to new individuals and situations, and, since aggregate fit does not guarantee individual-level validity~\citep{robinson2009ecological, zhang2026edit}, we evaluate it directly on individual behavior.

\xhdr{Theory of Mind Modeling} 
Existing Theory of Mind (ToM) frameworks and benchmarks heavily rely on synthetic narratives or explicit human annotations~\citep{gandhi2023understanding,kim2023fantom,wilf2024think,sclar2023minding,zhou2025social}, sparking debate over whether LLMs perform robust causal reasoning or superficial pattern matching~\citep{kosinski2024evaluating,ullman2023large,sap2022neural}. Our constituent-level decomposition directly operationalizes ToM by inferring how distinct groups update latent beliefs, but it induces this capability from macroscopic market outcomes. By bypassing micro-level annotations entirely, we learn grounded mental-state reasoning that transfers zero-shot to concrete individuals.

\vspace{-1mm}
\section{Problem Formulation}
\label{sec:problem}
\vspace{-1mm}

We formalize two related prediction tasks. The \emph{macro task} predicts how a population-level signal evolves in response to real-world events. The \emph{micro task} predicts how a specific individual responds to a given situation. We structure both tasks under a unified framework of \emph{observation}, \emph{policy}, and \emph{evaluation}. The observation captures the context and history available at a given step. The policy $\pi_\theta$ maps the observation to a reasoning trace followed by a prediction of the next outcome. 

Because both tasks share this input--output interface, a single policy $\pi_\theta$ can be trained only on the macro task and evaluated zero-shot on the micro task, with individual human responses reserved for testing. This setting directly tests our core question: can behavioral reasoning learned from macroscopic outcomes transfer to simulating individuals?

\subsection{Macro Task: Population-level Forecasting}
\label{sec:source}

\xhdr{Observation}
Each instance centers on a market question, a natural-language proposition such as the outcome of an election. Its price $y_t \in [0,1]$ reflects the collective decisions of market participants at step $t$. The model observes the question, the recent price history, and a set of text descriptions of real-world events, such as news items. We denote this macro observation by $x_t$.

\xhdr{Policy}
Given $x_t$, the policy generates a reasoning trace $\tau_t$ followed by a forecast of the next price, $(\tau_t, \hat{y}_{t+1}) \sim \pi_\theta(\cdot \mid x_t)$.

\xhdr{Evaluation}
The macro dataset collects transitions across many markets, $\mathcal{D}_{\mathrm{macro}} = \{(x_t, y_{t+1})\}$. The task is evaluated by how accurately the forecast $\hat{y}_{t+1}$ matches the realized next price $y_{t+1}$, using MAE, MASE, directional accuracy, and Pearson correlation.
\vspace{-1mm}
\subsection{Micro Task: Individual-level Simulation}
\label{sec:target}

\xhdr{Observation}
Each instance targets a specific user. At step $t$, the model observes the user's persona and interaction history $h_t$, such as a profile, stated views, or past messages, together with the current context, such as a news event, an ongoing conversation, or a candidate reply to be judged. We denote the full micro observation by $x_t$; all of it is available before the target response.

\xhdr{Policy}
Given $x_t$, the policy likewise generates a reasoning trace $\tau_t$ followed by a prediction of the user's response, $(\tau_t, \hat{y}_{t+1}) \sim \pi_\theta(\cdot \mid x_t)$.

\xhdr{Evaluation}
The micro dataset collects individual interactions, $\mathcal{D}_{\mathrm{micro}} = \{(x_t, y_{t+1})\}$, where $y_{t+1}$ is the user's actual response. The task is evaluated by how well $\hat{y}_{t+1}$ matches $y_{t+1}$, using task-specific scores such as rubric-based alignment for free-text replies, or error and rank correlation for ratings.

\section{Learning to Simulate Individuals from Macro Signals}
\label{sec:method}

\begin{figure}[t]
\vspace{-2mm}
\centering
\includegraphics[width=0.98\textwidth]{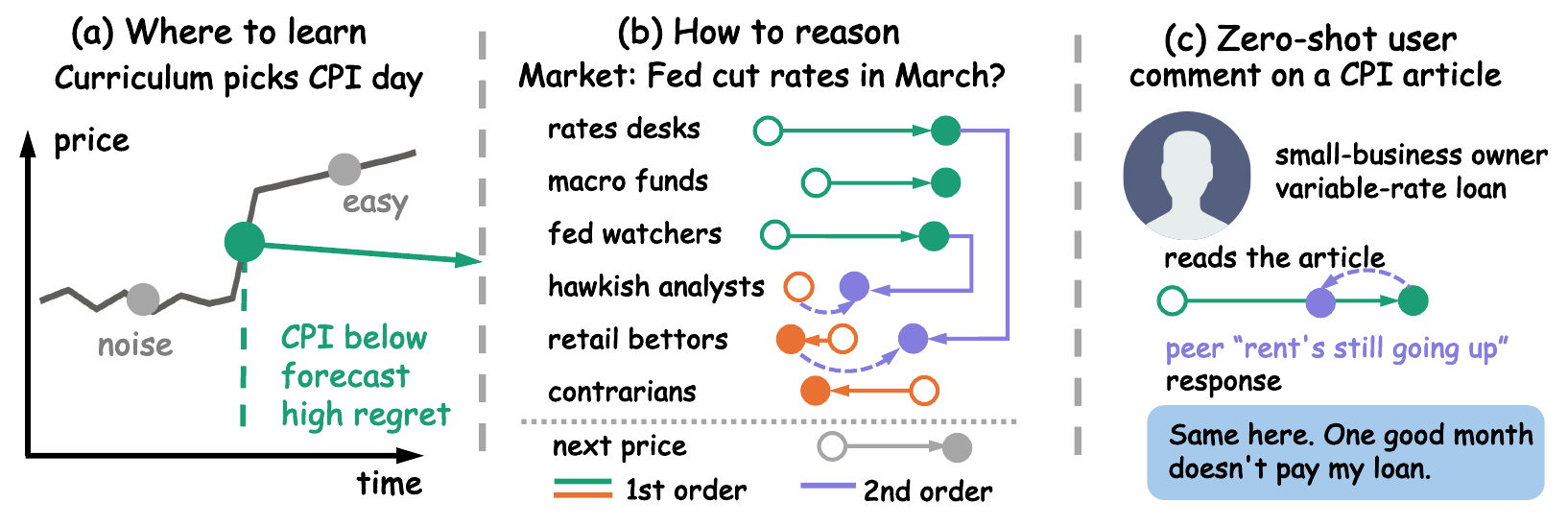}
\caption{
\textbf{The pipeline of \modelname, illustrated with a running example.} 
(a) \textbf{Where to learn:} The regret-and-difficulty curriculum performs data selection by focusing on high-regret transitions, such as the price jump after inflation data came in lower than expected, while ignoring the noise. 
(b) \textbf{How to reason:} Through social behavioral decomposition, the model infers how different groups react to the news (first order) and to one another (second order), and aggregates their belief updates into a forecast rewarded by the realized price.
(c) \textbf{Zero-shot transfer:} This learned reasoning ability is transferred zero-shot to user simulations, effectively modeling how an individual (e.g., a small-business owner) reacts to the CPI news and peer comments.
}
\vspace{-2mm}
\label{fig:method}
\end{figure}

We realize our approach through a single autoregressive language-model policy $\pi_\theta$. We first describe how the policy structures social behavioral reasoning during forecasting (Section~\ref{sec:structure}), how it is optimized via GRPO using a hindsight-regret curriculum (Section~\ref{sec:curriculum}), and how this learned reasoning transfers to individual simulation (Section~\ref{sec:transfer}).
\vspace{-2mm}
\subsection{Social Behavioral Reasoning}
\label{sec:structure}

\xhdr{Social behavioral decomposition}
We instantiate this reasoning as a single autoregressive trajectory, avoiding the cost and credit-assignment difficulties of a full multi-agent simulation~\citep{yang2024oasis}. Given an observation $x$, the policy generates a reasoning trace $\tau = (\tau^{\mathrm{prop}}, \tau^{F}, \tau^{A})$ before outputting the forecast $\hat{y}$. The resulting decomposition has natural counterparts in individual simulation.

\textbf{1. Constituent initialization ($\tau^{\mathrm{prop}}$).} The policy identifies representative constituents likely to drive the market shift, and constructs a prior state for each that describes its role, objectives, and beliefs. \textit{Micro counterpart:} inferring which constituents drive a given change, and what they care about, requires linking behavioral patterns to the kinds of people who produce them. This is the same persona--behavior association that individual simulation relies on.

\textbf{2. First-order responses ($\tau^{F}$).} The policy predicts how each constituent updates its beliefs and responds based on its own prior and the observation. \textit{Micro counterpart:} predicting a user's belief update and immediate response to the current context.

\textbf{3. Second-order responses and aggregation ($\tau^{A}$).} The policy models the most consequential interactions, assessing how one constituent's reaction alters another's beliefs and actions, and then aggregates the adjusted responses into the forecast $\hat{y}$. \textit{Micro counterpart:} adapting to others' reactions, as in replying within a multi-turn dialogue.

\xhdr{Behavioral reasoning as a shared transition function}
Both tasks rest on a shared behavioral transition function, which maps an agent's prior state and context to a belief update and an observable response, and can be viewed as a computational form of theory of mind. The two tasks differ in the observability of the prior state. In the micro task, it is given by the user's history. In the macro task, it is unobserved, and the policy must first infer the prior states of key constituents before predicting their responses. Because the behavioral
transition function is shared across constituents and tasks, the behavioral reasoning ability that the policy learns from macro rewards can be applied to individual simulation without further training (Section~\ref{sec:transfer}).

\xhdr{Orders of social behavioral reasoning}
Social reactions to an event unfold in orders. At the first order, each group reacts directly to the news; at the second order, groups react to one another's reactions, and such cascades can in principle continue further. Individuals behave the same way, responding both to events and to what others say about them. Our template reasons up to the second order and focuses on the few interactions that matter most, which captures effects such as persuasion and conformity while maintaining a concise autoregressive trajectory. This way of reasoning bounds the intermediate reasoning process without altering the final macroscopic prediction target.

\subsection{Learning Social Behavioral Reasoning from Macro Signals}
\label{sec:curriculum}

\begin{algorithm}[h]
\footnotesize
\caption{Overview of \modelname training with the hindsight-regret curriculum and difficulty-aware sampling}
\label{alg:training}
\begin{algorithmic}[1]
\Require Dataset $\mathcal{D}_{\mathrm{macro}}$; initial policy $\pi_\theta$; reward function $R$; group size $M$; probe pool size $P$; difficulty filter $d$; weights $\lambda_{\mathrm{reg}}, \lambda_{\mathrm{diff}}$
\Ensure Trained parameters $\theta$

\State Initialize $w(x) \gets 1$ for all $x \in \mathcal{D}_{\mathrm{macro}}$

\For{each training step}

    \If{periodic probe} \Comment{refresh curriculum statistics}
        \State Sample a pool $\mathcal{P}$ of $P$ transitions uniformly from $\mathcal{D}_{\mathrm{macro}}$
        \For{each $(x, y) \in \mathcal{P}$}
            \State $c \sim \pi_\theta^{\mathrm{hint}}(\cdot \mid x, y)$ \Comment{hindsight hint}
            \State $(\tau_i, \hat{y}_i) \sim \pi_\theta(\cdot \mid x)$, \;
            $(\tau_i^{\mathrm H}, \hat{y}_i^{\mathrm H}) \sim
            \pi_\theta(\cdot \mid x, c)$, \;
            $i=1,\dots,M$

            \State $\Delta(x) \gets
            \frac{1}{M}\sum_{i=1}^{M}\tilde{R}(\hat{y}_i^{\mathrm H})
            -
            \frac{1}{M}\sum_{i=1}^{M}\tilde{R}(\hat{y}_i)$

            \State $p(x) \gets
            \frac{1}{M}\sum_{i=1}^{M}
            \mathbbm{1}\!\left[\tilde{R}(\hat{y}_i)\ge 0.5\right]$

            \State $w(x) \gets
            \lambda_{\mathrm{reg}}\max(\Delta(x),0)
            +
            \lambda_{\mathrm{diff}}\,d(p(x))$
        \EndFor
    \EndIf

    \State $\mathcal{B} \sim \mathcal{D}_{\mathrm{macro}} \propto w(x)$
    \Comment{sample training batch}

    \For{each $(x,y)\in\mathcal{B}$}
        \State $(\tau_i,\hat{y}_i)\sim\pi_\theta(\cdot\mid x)$,
        $i=1,\dots,M$

        \State $p(x)\gets
        \frac{1}{M}\sum_{i=1}^{M}
        \mathbbm{1}\!\left[\tilde{R}(\hat{y}_i)\ge0.5\right]$, $w(x) \gets
        \lambda_{\mathrm{reg}}\max(\Delta(x)_{\text{cached}},0)
        +
        \lambda_{\mathrm{diff}}\,d(p(x))$
    \EndFor

    \State $\theta \gets
    \textsc{GRPO-Update}(\theta,\mathcal{B},R)$
    \Comment{group-normalized advantages}
\EndFor

\State \Return $\theta$
\end{algorithmic}
\end{algorithm}

Macro signals are inherently noisy and uneven in difficulty (\textbf{C2}): many transitions are dominated by random fluctuations, some are trivially solvable, and only some actually reward reasoning about the latent constituents driving the market. Because GRPO gains minimal signal from transitions the model always or never solves, we augment policy optimization with a hindsight-regret curriculum that combines behavioral informativeness with current learning difficulty. Hindsight regret prioritizes transitions where the model overlooks key constituents, while a difficulty-aware filter favors examples that have informative within-group variation for GRPO.

\xhdr{Policy optimization}
For a transition $\in \mathcal{D}_{\mathrm{macro}}$, the policy samples a group of $M$ outputs $(\tau_i, \hat{y}_i) \sim \pi_\theta(\cdot \mid x)$, $i = 1, \dots, M$, each consisting of a reasoning trace $\tau_i$ and a forecast $\hat{y}_i$. To bootstrap the structured reasoning format and stabilize subsequent reinforcement learning~\citep{guo2025deepseek,li2026ava}, we first perform a lightweight supervised fine-tuning (SFT) warmup on a small seed set of structured behavioral traces. For a given market observation $x$, the policy samples a group of $M$ reasoning and forecasting trajectories. Each output is evaluated by a reward $R$ that combines formatting, numerical accuracy, and directional accuracy. We optimize the policy via GRPO~\citep{shao2024deepseekmath} using group-normalized advantages. To isolate forecasting capability, the curriculum below relies on a format-free reward function $\tilde{R}$, which drops the format term and renormalizes the remaining terms, so that it measures forecast quality alone.

\xhdr{Hindsight-regret curriculum}
We formulate the curriculum as a dynamic data sampler. Periodically, we probe a random subset of transitions to estimate their learning value. For a given transition, the current policy is provided the true outcome $y$ via a dedicated hint prompt and tasked with identifying the key constituents that drove the market. This generates a hint $c$ containing constituent names and their qualitative price directions (up or down only).

We inject $c$ into the standard prompt as a simulated analyst's insight, and sample $M$ hinted outputs $(\tau^{\mathrm{H}}_i, \hat{y}^{\mathrm{H}}_i) \sim \pi_\theta(\cdot \mid x, c)$ alongside $M$ unhinted outputs $(\tau_i, \hat{y}_i) \sim \pi_\theta(\cdot \mid x)$. We compute the hindsight regret $\Delta$ and the baseline win rate $p$, estimating both from these samples:
\begin{equation}
\small
\Delta = \mathbb{E}\big[\tilde{R}(\hat{y}^{\mathrm{H}})\big] - \mathbb{E}\big[\tilde{R}(\hat{y})\big], \; p = \Pr\big(\tilde{R}(\hat{y}) \ge 0.5\big).
\label{eq:regret}
\end{equation}
Here, $\Delta$ isolates the capability gap: what the policy can successfully explain when aware of the key constituents, yet fails to anticipate without them. Because $c$ is generated by the policy itself, this gap shifts as the model improves.

\xhdr{Sampling weights}
We combine these two statistics into a composite sampling weight, adapting the designer reward formulation from SPADE~\citep{liu2026spade}:
\begin{equation}
w = \lambda_{\mathrm{reg}} \max(\Delta, 0) + \lambda_{\mathrm{diff}}\, d(p),
\label{eq:weight}
\end{equation}
where $\lambda_{\mathrm{reg}}$ and $\lambda_{\mathrm{diff}}$ control the relative importance of each component. These terms serve complementary purposes: the regret term evaluates whether a transition is \emph{informative}, while the difficulty term determines whether it is \emph{currently learnable}.

Specifically, the regret term prioritizes transitions whose outcomes hinge on critical constituents that the policy currently overlooks. We floor this value at zero because a negative regret implies that the self-generated hint misled the policy, rather than exposing a genuine capability gap. The difficulty term accounts for the optimization dynamics of GRPO. Because GRPO learns from the relative differences among $M$ responses to the same transition, the advantage signals are strongest when a group contains a mix of successes and failures. This variance peaks at intermediate success rates and vanishes when the policy consistently succeeds or fails, regardless of the regret magnitude. Consequently, we define $d(p)$ as a trapezoidal filter that equals 1 for $p \in [p_{\mathrm{lo}}, p_{\mathrm{hi}}]$ and decays linearly to 0 as $p \to 0$ or $p \to 1$.

Together, the two terms favor transitions that are both behaviorally informative and within the policy's current learning capacity. Transitions not yet probed receive the mean weight of the probed ones, and the win rate $p$ is also updated during regular GRPO steps. Each training batch $\mathcal{B}$ is sampled from $\mathcal{D}_{\mathrm{macro}}$ with probability proportional to $w$. Algorithm~\ref{alg:training} gives an overview of the training procedure.

\subsection{Applying Social Behavior Reasoning to Individual Simulation}
\label{sec:transfer}

The core premise of our approach is that macroscopic market shifts and microscopic user actions are driven by the same underlying behavioral mechanics. Consequently, we transfer the reasoning capabilities learned from macro signals directly to individual simulation. 

\xhdr{Individual-level social reasoning}
For individual simulation, the prior state is given rather than inferred. Depending on the task, it comes from the user's interaction history, such as past posts or conversations, or from a persona, such as a profile and stated views. We therefore bypass constituent discovery and directly instantiate the prior state of the target individual with this information. Given this prior state and the current context, the policy applies the same learned behavioral transition to infer how the user interprets the context, updates their beliefs, and responds.

\xhdr{Inference}
We apply the trained policy to individual simulation zero-shot, without any parameter updates, and evaluate only the final response. During training, the policy reasons about coarse groups inferred from market context; here, it applies the same behavioral reasoning process to a specific person with an observed history or persona.

\xhdr{Synthetic data}
The same procedure can also generate training data to address the scarcity of individual behavioral data. Given a user's persona and a few-shot history, optionally augmented with population-level behavioral statistics available in the downstream setting, we apply the trained policy to generate both novel questions and the corresponding responses. This synthetic data is then used to fine-tune and align a separate downstream simulator.

\section{Main Experiments}
\label{sec:exp_setup}

\begin{table*}[t]
\centering
\caption{
\textbf{Results on SWM-Bench.}
\textbf{Bold} / \underline{underline}: best / second best results.
}
\label{tab:swmbench}
\scriptsize
\setlength{\tabcolsep}{2pt}
\renewcommand{\arraystretch}{0.90}

\resizebox{0.92\textwidth}{!}{%
\begin{tabular}{l rrrr rrrr}
\toprule
\multirow{2}{*}{\textbf{Method}}
& \multicolumn{4}{c}{\textbf{Polymarket}}
& \multicolumn{4}{c}{\textbf{Kalshi}} \\
\cmidrule(lr){2-5}\cmidrule(lr){6-9}
& MASE$\downarrow$ & MAE$\downarrow$ & DA$\uparrow$ & Corr$\uparrow$
& MASE$\downarrow$ & MAE$\downarrow$ & DA$\uparrow$ & Corr$\uparrow$ \\
\midrule

\multicolumn{9}{l}{\cellcolor{rowgray}\textit{Time-series only}} \\
DLinear~\citep{zeng2023transformers}
  & 1.111 & 0.048 & 0.580 & \underline{0.348}
  & 1.245 & 0.075 & 0.492 & $-$0.035 \\
TimeMixer~\citep{wang2024timemixer}
  & 0.997 & \underline{0.043} & 0.590 & 0.224
  & 1.079 & 0.065 & 0.549 & $-$0.056 \\
PatchTST~\citep{nie2022time}
  & 0.994 & \underline{0.043} & \underline{0.592} & 0.306
  & 1.174 & 0.071 & 0.536 & $-$0.035 \\

\midrule
\multicolumn{9}{l}{\cellcolor{rowgray}\textit{Time-series + news (Prompting-based LLM)}} \\
Qwen3-4B~\citep{yang2025qwen3}
  & 2.297 & 0.099 & 0.584 & 0.088
  & 1.848 & 0.112 & 0.559 & 0.098 \\
Qwen3-4B-think~\citep{yang2025qwen3}
  & 2.602 & 0.112 & 0.511 & 0.075
  & 1.951 & 0.118 & 0.529 & 0.030 \\
GPT-5.5~\citep{openai2026gpt55}
  & \bfseries 0.985 & \bfseries 0.042 & 0.337 & 0.264
  & \bfseries 0.997 & \bfseries 0.060 & 0.564 & \bfseries 0.242 \\

\midrule
\multicolumn{9}{l}{\cellcolor{rowgray}\textit{Time-series + news (Fine-tuned LLM)}} \\
FNF~\citep{wang2024newsforecastintegratingevent}
  & \underline{0.986} & \bfseries 0.042 & 0.377 & 0.218
  & 1.126 & 0.068 & 0.430 & $-$0.020 \\
LLM4TS~\citep{chang2025llm4ts}
  & 0.993 & \underline{0.043} & 0.589 & 0.312
  & 1.125 & 0.068 & 0.554 & $-$0.015 \\
SWM~\citep{yu2026building}
  & 1.141 & 0.049 & 0.535 & 0.088
  & 1.013 & \underline{0.061} & \bfseries 0.589 & \underline{0.167} \\
GRPO-8B
  & 1.044 & 0.045 & 0.475 & 0.012
  & \underline{1.001} & \bfseries 0.060 & 0.448 & 0.031 \\

\midrule
\rowcolor{rowblue}\textbf{macro2mind(4B)}
  & 0.999 & \underline{ 0.043} & \bfseries 0.674 & \bfseries 0.362
  & 1.088 & 0.066 & \underline{0.569} & 0.016 \\

\bottomrule
\end{tabular}%
}
\end{table*}

\subsection{Problem Setup}
\xhdr{Market prediction}
We train and evaluate our model on \textbf{SWM-Bench}~\citep{yu2026building}, which pairs prediction-market trajectories with contemporaneous news. The task is to predict the next observed market price, providing aggregate supervision without requiring any annotations of the underlying individual behaviors.
\xhdr{User Simulation}
To evaluate the zero-shot transfer of behavioral reasoning across both single-turn and multi-turn interactions, we select four benchmarks spanning three distinct individual-level tasks: free-text response generation (\textbf{Humanual}~\citep{wu2026humanlm}, Qwen3.8-Max as judge), representation judgment (\textbf{OvertonBench}~\citep{poole2026benchmarking}), and conversational preference prediction (\textbf{PRISM}~\citep{kirk2024prism}\footnote{Following the original paper, we multiply the prompted 0--1 scores by 100 to report results on a 0--100 scale.} and \textbf{CAD}~\citep{zhang2026cultivating}).

\xhdr{Synthetic data}
We evaluate on \textbf{Twin-2K-500}~\citep{toubia2025database}.
Each generator writes new scenarios for 600 training users and labels them by answering as each participant, conditioned on the persona and the answer distribution of training users.
A Qwen3-4B student is fine-tuned on each synthetic set and evaluated on real held-out questions, either for 100 unseen users or for the source users themselves. Details are in Appendix~\ref{app:synthetic}.

\subsection{Baseline settings.}
\xhdr{Market Prediction} We compare \modelname against three categories of methods. (\textbf{1}) \emph{Time-series only} baselines forecast solely from the numerical historical window and are trained on SWM-Bench: DLinear, PatchTST and TimeMixer. (\textbf{2}) \emph{Prompting-based LLMs} process both the numerical window and all available news to generate forecasts via zero-shot prompting without parameter updates: Qwen3-4B, GPT-5.5. (\textbf{3}) \emph{Fine-tuned LLMs} also use both modalities but are trained directly on SWM-Bench. SWM, FNF, LLM4TS and GRPO all use Qwen3-8B. \modelname uses Qwen3-4B as backbone.

\xhdr{User Simulation}
We compare \modelname against four categories of methods.
(\textbf{1}) \emph{Same-backbone baselines}: zero-shot Qwen3-4B and standard GRPO trained on SWM-Bench. 
(\textbf{2}) \emph{Prompting-based simulators}: training-free inference methods, including Qwen3-4B, PERSONA-Flow, SynthesizeMe, Self-consistency, History-ICL and Self-report. 
(\textbf{3}) \emph{Trained user simulators}: fine-tuned models including HumanLM-SFT, -SFT-think, and -GRPO. 
(\textbf{4}) \emph{Personalized reward models}, repurposed as simulators: the generative P-GenRM, alongside the scalar models Skywork-RM, MRM, and WIMHF-RM. Scalar reward models generate no text and instead rerank eight Qwen3-4B samples. All methods use Qwen3-4B except Skywork-RM, and MRM. Details in Appendix.~\ref{app:usersim_baselines}.
\begin{table}[t]
\centering
\caption{
\textbf{Results on user simulation tasks.}
Parentheses denote training data (\emph{per bench.}: one model trained per benchmark).
\textbf{Bold} / \underline{underline}: best / second best among zero-shot (out-of-domain) results. We multiply all Humanual scores by 100 for reporting.
}
\label{tab:user_simulation}
\small
\setlength{\tabcolsep}{2.5pt}
\renewcommand{\arraystretch}{1}
\begin{tabular}{@{}l ccccccc@{}}
\toprule
& \multicolumn{2}{c}{Humanual}
& \multicolumn{2}{c}{Overton}
& \multicolumn{2}{c}{PRISM}
& CAD \\
\cmidrule(lr){2-3}\cmidrule(lr){4-5}\cmidrule(lr){6-7}\cmidrule(lr){8-8}
Method
& State$\uparrow$ & Resp.$\uparrow$
& MAE$\downarrow$ & $\rho\uparrow$
& MAE$\downarrow$ & Acc.$\uparrow$
& Acc.$\uparrow$ \\
\midrule
\multicolumn{8}{@{}l} {\cellcolor{rowgray}{\textit{Trained user simulators}}} \\
HumanLM-SFT {\scriptsize(Humanual)}~\citep{wu2026humanlm}       & 8.2 & 3.2 & 0.734 & 0.566 & 22.5 & 59.6 & 35.0 \\
HumanLM-SFT-think {\scriptsize(Humanual)} & 6.3 & 1.8 & 0.754 & 0.606 & 23.3 & 60.1 & 36.5 \\
HumanLM-GRPO {\scriptsize(Humanual)}      & 12.4 & 5.3 & 0.718 & 0.576 & 21.8 & 62.6 & 39.1 \\
\midrule
\multicolumn{8}{@{}l}{\cellcolor{rowgray}{\textit{Personalized reward models}}} \\
P-GenRM {\scriptsize(PRISM)}~\citep{zhang2026p}       & 3.4 & 0.7 & 0.786 & 0.455 & 21.8 & 66.2 & \textbf{44.0} \\
Skywork-RM {\scriptsize(pretrained)}~\citep{liu2026skywork} & 8.3 & 2.4 & 1.024 & 0.090 & 21.5 & \textbf{64.7} & 41.8 \\
MRM {\scriptsize(per bench.)}~\citep{cai2026one}        & 7.8 & 2.5 & 0.907 & 0.338 & 20.7 & 69.1 & 44.2 \\
WIMHF-RM {\scriptsize(per bench.)}~\citep{movva2026s}   & 7.9 & 2.6 & 0.859 & 0.429 & 20.7 & 69.5 & 43.6 \\
\midrule
\multicolumn{8}{@{}l}{\cellcolor{rowgray}{\textit{Prompting-based simulators}}} \\
PERSONA-Flow~\citep{feng2026persona}                  & 7.5 & 2.1 & 0.756 & 0.540 & 21.5 & 62.5 & 38.8 \\
SynthesizeMe ~\citep{ryan2025synthesizeme}                  & 7.7 & 2.2 & 0.763 & 0.529 & 21.3 & 63.2 & 40.4 \\
Self-cons.\ ($n{=}5$) ~\citep{wang2022self}         & 7.9 & 2.1 & 0.758 & 0.538 & 21.3 & 62.6 & 38.6 \\
History-ICL  ~\citep{hwang2023aligning}                  & \underline{8.8} & \underline{2.9} & 0.741 & 0.552 & \textbf{20.5} & \underline{63.6} & \underline{42.9} \\
Self-report  ~\citep{park2024llm}                  & 6.9 & 1.8 & 0.750 & 0.544 & 21.5 & 62.7 & 38.5 \\
\midrule
\multicolumn{8}{@{}l}{\cellcolor{rowgray}{\textit{Same-backbone controls}}} \\
Qwen3-4B {\scriptsize(zero-shot)}  & 7.7 & 2.4 & 0.756 & 0.539 & 21.5 & 62.4 & 38.9 \\
GRPO-4B {\scriptsize(SWMBench, zero-shot)}       & 6.1 & 1.5 & \textbf{0.623} & \underline{0.647} & 23.0 & 53.9 & 36.6 \\
\midrule
\rowcolor{rowblue}\textbf{\modelname} {~{\scriptsize(SWMBench, zero-shot)}} & \textbf{9.5} & \textbf{3.4} & \underline{0.635} & \textbf{0.653} & \underline{21.2} & 63.3 & 42.2 \\
\bottomrule
\end{tabular}
\end{table}

\subsection{Evaluation settings.}
For \textbf{Market Prediction}, we evaluate Mean Absolute Scaled Error (MASE), Mean Absolute Error (MAE), Directional Accuracy (DA) and Pearson Correlation (Corr). For \textbf{User Simulation}, we evaluate predictions against ground-truth behaviors. We report {state} and {response alignment} via LLM-as-a-judge on {Humanual-Politics}; MAE and correlations $\rho$ on {OvertonBench}; MAE and preference accuracy on PRISM. Preference accuracy on {CAD}. Details in Appendix~\ref{app:usersim_metrics}.

\subsection{Main Results}

\textbf{\modelname improves market forecasting, particularly in predicting directional movement.}
On SWM-Bench (Table~\ref{tab:swmbench}), \modelname achieves the highest directional accuracy (0.674) and correlation (0.362) on Polymarket---surpassing both the strongest time-series baseline and GPT-5.5---while ranking second in directional accuracy on Kalshi (0.569). 

\textbf{\modelname demonstrates robust zero-shot transfer.}
Trained on prediction market signals, \modelname has competitive performance among zero-shot methods (Table~\ref{tab:user_simulation}). It secures the best Humanual alignment (9.5/3.4) and OvertonBench $\rho$ (0.653), remarkably surpassing two supervised HumanLM variants zero-shot. Furthermore, it firmly out-generalizes domain-specific simulators: \modelname beats Humanual-trained HumanLM-GRPO across all five out-of-domain metrics (e.g., CAD: 42.2\% vs.\ 39.1\%) and avoids PRISM-trained P-GenRM's catastrophic collapse on Humanual. On PRISM and CAD accuracy, \modelname remains highly competitive (63.3\% / 42.2\%), trailing only History-ICL and dataset-specific reward models.

\textbf{\modelname serves as a highly effective synthetic data generator for downstream user simulators.}
Fine-tuning a student model on data generated by \modelname boosts its accuracy from 44.5\% to 60.0\% on unseen users (out-of-user) and from 50.9\% to 61.3\% on source users (in-user). In contrast, data generated by its Qwen3-4B backbone using the identical pipeline yields only marginal improvements (46.8\% and 53.0\%, respectively). Since the two generators differ exclusively in their exposure to macro market forecasting, this 13.2-point gap on unseen users demonstrates that \modelname is a good teacher for micro-level simulation.

\section{Discussion}
\begin{figure*}[t]
\centering
\vspace{-3mm}
\begin{minipage}[t]{0.32\linewidth}
\vspace{0pt}
\centering
\includegraphics[width=\linewidth]{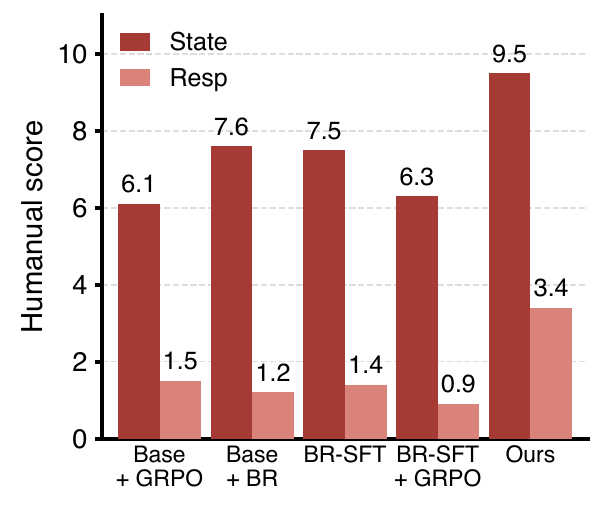}
\caption{\textbf{Component ablation results on Humanual.} We use Qwen3-4B as base model.}
\label{fig:rq1}
\end{minipage}
\hfill
\begin{minipage}[t]{0.32\linewidth}
\vspace{0pt}
\centering
\includegraphics[width=\linewidth]{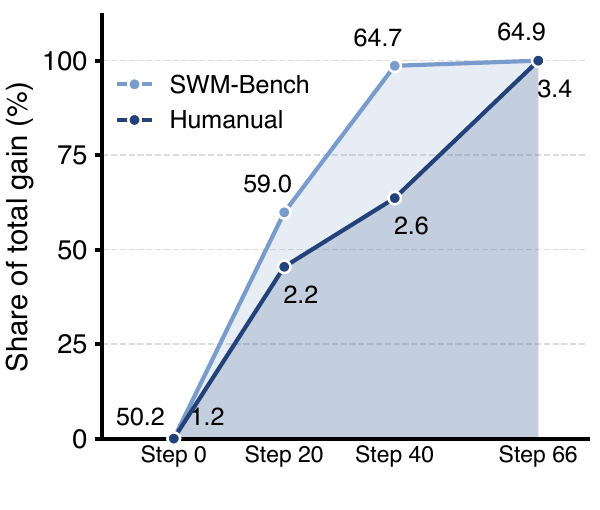}

\caption{\textbf{Training dynamics.}  SWM-Bench DA and Humanual Response score across training checkpoints.}
\label{fig:co_improve}
\end{minipage}
\hfill
\begin{minipage}[t]{0.32\linewidth}
\vspace{0pt}
\centering
\includegraphics[width=\linewidth]{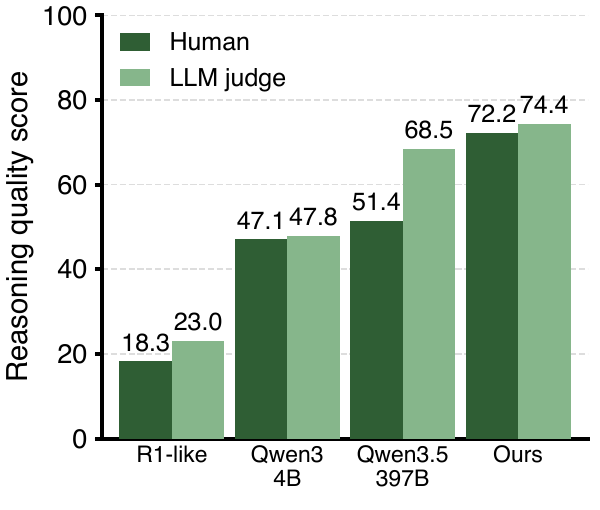}

\caption{\textbf{Human evaluation of reasoning plausibility.} Pearson $r = 0.885$.}
\label{fig:human-eval}
\end{minipage}
\vspace{-6mm}
\end{figure*}

\textbf{RQ1: Which part of the training recipe drives transfer?}
\label{sec:rq1}
\textit{Ablation results show that the full curriculum, combining hindsight regret with difficulty-aware sampling, is the key driver of zero-shot transfer.} To isolate their contributions, we evaluate additive ablations on Humanual: {Base+BR} (zero-shot prompting), {BR-SFT} (fine-tuning on BR traces), {BR-SFT+GRPO} (GRPO without curriculum), {Base+GRPO} (GRPO with free form thinking) and \modelname (full curriculum) under identical hyperparameters. Results (Figure~\ref{fig:rq1}) show the BR structure alone cannot explain the gains: {Base+BR} and {BR-SFT} yield similar State scores (7.6 vs.\ 7.5). Neither BR-SFT nor BR-SFT+GRPO improves over Base+BR: BR-SFT stays at 7.5 State and 1.4 Resp, and adding GRPO without the curriculum gives 6.3 State and 0.9 Resp. Thus, naively optimizing for market forecasting does not inherently transfer to individual simulation. Meaningful transfer emerges only when adding the full regret-and-difficulty curriculum: \modelname reaches 9.5 State and 3.4 Resp, improving over Base+BR by 25\% in State score. 

\paragraph{RQ2: Do macro forecasting and micro simulation ability improve together during training?}
\label{sec:rq2}

\textit{Results indicate a concurrent improvement during training. As \modelname learns to forecast macro signals, its zero-shot user-simulation capability exhibits corresponding gains.} Evaluating four training checkpoints (steps 0, 20, 40, and 66) on both SWM-Bench and Humanual (Figure~\ref{fig:co_improve}) reveals a striking trajectory. Over the first 40 steps, both abilities surge simultaneously—SWM-Bench directional accuracy increases from 50.2\% to 64.7\%, while Humanual response alignment more than doubles (1.2 $\rightarrow$ 2.6) despite zero exposure to its data. Beyond step 40, forecasting accuracy plateaus (64.9\%), yet response alignment continues to climb to nearly three times its initial value (3.4). This continuous co-improvement demonstrates that zero-shot transfer is driven by the acquisition of macro behavioral reasoning.

\paragraph{RQ3: Does the learned behavioral reasoning align with human judgments?}
\label{sec:rq3}

\textit{Human evaluation indicates that the behavioral reasoning generated by \modelname aligns well with human judgments.} Five annotators evaluated 20 traces per model (0--100 scale), judging whether the inferred groups, news interpretations, and belief updates coherently explained the final forecast. As shown in Figure~\ref{fig:human-eval}, \modelname scores 72.2, surpassing both Qwen3-4B (47.1) and Qwen3.5-397B (51.4) under the exact same structural prompt. This suggests that structural prompting alone is insufficient to achieve human-aligned reasoning. In contrast, an R1-style model reasoning freely within \texttt{<think>} tags scores 18.3, often defaulting to disjointed numerical speculation rather than modeling actual human perspectives. These human ratings correlate strongly with our LLM judge (Pearson $r = 0.885$).

\vspace{-1mm}
\section{Conclusion}
\label{sec:conclusion}
\vspace{-1mm}

We introduced \modelname, which learns to simulate individual human behavior from aggregate prediction-market signals via a structured social behavioral decomposition and a dynamic hindsight-regret curriculum. \modelname achieves state-of-the-art DA and Corr on Polymarket while demonstrating robust zero-shot transfer to user simulation. Furthermore, its generated data can serve as high-quality supervision, significantly boosting the accuracy of downstream models. In the future, exploring other forms of macroscopic supervision and deploying these simulators into dynamic, multi-agent ecosystems remain exciting directions.

\subsection*{AI use statement}
AI systems, including LLMs, did not contribute to the development of the research ideas or the writing of this paper to a degree that would warrant authorship or contributor status. In this work, LLMs are treated strictly as objects of study and are the primary focus of our experimental evaluations.

\section*{Ethics Statement}
This work introduces \modelname, it is a methodological contribution rather than a deployed system, and its societal impact is largely mediated by how downstream practitioners apply it. All experiments are conducted on standard, publicly available benchmarks and open-weight or proprietary models used strictly for research evaluation; the work does not involve the collection of new personal data.

\subsection*{Reproducibility statement}
We describe the full training and evaluation setup needed to reproduce our main results. Evaluation metrics are formally defined in Appendix~\S\ref{appendix:metrics}. We will release our training code and configurations upon acceptance.

\bibliography{iclr2027_conference}
\bibliographystyle{iclr2027_conference}

\appendix
\section{Limitation}

Our work has several limitations that suggest directions for future research. First, due to computational constraints, almost all experiments use Qwen3 as the shared backbone. Scaling \modelname to larger models (e.g., 32B–70B) and validating its behavior across other model families (e.g., LLaMA-3) is an important next step. Second, due to the limitation of time and resources, we did not conduct large-scale human evaluation; we will conduct large-scale human evaluation in future research.

\section{Reward Function of \modelname Training}
\label{app:reward}

For a generated output $o$ on transition $t$, containing a forecast
$\hat{y}_{t+1}$ for the realized next price $y_{t+1}$, the training reward
is a weighted sum of three terms:
\begin{equation}
R(o)
=
\alpha_{\mathrm{fmt}} R_{\mathrm{fmt}}(o)
+
\alpha_{\mathrm{corr}} R_{\mathrm{corr}}(o)
+
\alpha_{\mathrm{dir}} R_{\mathrm{dir}}(o).
\label{eq:reward_full}
\end{equation}

\xhdr{Format}
$R_{\mathrm{fmt}}(o)=1$ if the output contains every segment of the
reasoning template and a parsable forecast $\hat{y}_{t+1}\in[0,1]$,
and $R_{\mathrm{fmt}}(o)=0$ otherwise. If no valid forecast can be
parsed, we set $R_{\mathrm{corr}}(o)=R_{\mathrm{dir}}(o)=0$.

\xhdr{Correctness}
For a valid forecast, the correctness reward decays exponentially with
the absolute forecast error:
\begin{equation}
R_{\mathrm{corr}}(o)
=
\exp\!\left(-\left|\hat{y}_{t+1}-y_{t+1}\right|\right).
\end{equation}
Since prices lie in $[0,1]$, this term lies in $[e^{-1},1]$ for valid
forecasts.

\xhdr{Direction}
The direction reward indicates whether the predicted price change has
the same sign as the realized change relative to the current price
$y_t$:
\begin{equation}
R_{\mathrm{dir}}(o)
=
\mathbbm{1}\!\left[
\operatorname{sign}(\hat{y}_{t+1}-y_t)
=
\operatorname{sign}(y_{t+1}-y_t)
\right].
\end{equation}

\xhdr{Format-free reward}
The curriculum (Section~\ref{sec:curriculum}) evaluates forecast quality
independently of output-format compliance. We therefore remove the format
term and renormalize the remaining weights:
\begin{equation}
\tilde{R}(o)
\equiv
R^{\setminus\mathrm{fmt}}(o)
=
\frac{
\alpha_{\mathrm{corr}}R_{\mathrm{corr}}(o)
+
\alpha_{\mathrm{dir}}R_{\mathrm{dir}}(o)
}{
\alpha_{\mathrm{corr}}+\alpha_{\mathrm{dir}}
}
\in[0,1].
\label{eq:reward_format_free}
\end{equation}

\providecommand{\modelname}{macro2mind}
\definecolor{promptframe}{HTML}{34546E}
\definecolor{promptback}{HTML}{F7F9FB}
\newtcolorbox{promptbox}[2][]{enhanced jigsaw,breakable,
  colback=promptback,colframe=promptframe,boxrule=0pt,leftrule=2.4pt,arc=0pt,outer arc=0pt,
  left=9pt,right=9pt,top=5pt,bottom=6pt,
  title={#2},title after break={},fonttitle=\small\bfseries\sffamily,
  coltitle=promptframe,colbacktitle=promptframe!10,toptitle=3pt,bottomtitle=3pt,lefttitle=9pt,
  pad at break*=1mm,
  fontupper=\small,before upper={\setlength{\parindent}{0pt}\setlength{\parskip}{2pt}\sloppy},
  segmentation style={dashed,draw=promptframe!40},
  before skip=12pt,after skip=12pt,#1}
\newcommand{\promptrole}[1]{\par\addvspace{3pt}\noindent{\sffamily\footnotesize\bfseries\color{promptframe}#1 prompt}\par\nobreak\vspace{1pt}}
\section{Evaluation Details}
\label{appendix:metrics}

\subsection{SWM-Bench}
Let the test set contain $N$ examples. For example $i$, let $p_i$ be the last observed price (at time $t$), $y_i$ the realized next price (at $t{+}1$), and $\hat{y}_i$ the predicted price.

\xhdr{Mean Absolute Error (MAE)} The average magnitude of the prediction error:
\begin{equation}
\mathrm{MAE} = \frac{1}{N}\sum_{i=1}^{N}\bigl|\hat{y}_i - y_i\bigr|.
\end{equation}

\xhdr{Mean Absolute Scaled Error (MASE)} MAE normalized by the error of the \emph{persistence} (no-change) baseline $\hat{y}_i = p_i$, which scales away cross-market differences in volatility; $\mathrm{MASE}<1$ means the model beats copying the last price:
\begin{equation}
\mathrm{MASE} = \frac{\sum_{i}\bigl|\hat{\Delta}_i - \Delta_i\bigr|}{\sum_{i}\bigl|\Delta_i\bigr|}.
\end{equation}

\xhdr{Three-way Directional Accuracy (DA)} We classify each change as \emph{up}, \emph{down}, or \emph{no-change} via $\mathrm{sgn}_\delta(x) \in \{+1, -1, 0\}$ with a small dead band $\delta = 10^{-6}$ ($+1$ if $x > \delta$, $-1$ if $x < -\delta$, $0$ otherwise), and measure agreement over the \emph{moved} subset $\mathcal{M}=\{i : |\Delta_i| > \delta\}$:
\begin{equation}
\mathrm{DA} = \frac{1}{|\mathcal{M}|}\sum_{i\in\mathcal{M}}
\mathbbm{1}\!\left[\mathrm{sgn}_\delta(\hat{\Delta}_i)=\mathrm{sgn}_\delta(\Delta_i)\right].
\end{equation}
The three-way scheme penalizes predicting ``no change'' on a market that genuinely moved.

\xhdr{Pearson Correlation (Corr)} The Pearson correlation between predicted and realized belief changes over all $N$ examples, capturing how well the model tracks the magnitude and sign of moves:
\begin{equation}
\mathrm{Corr} =
\frac{\sum_{i}(\hat{\Delta}_i-\bar{\hat{\Delta}})(\Delta_i-\bar{\Delta})}
     {\sqrt{\sum_{i}(\hat{\Delta}_i-\bar{\hat{\Delta}})^2}\,
      \sqrt{\sum_{i}(\Delta_i-\bar{\Delta})^2}},
\end{equation}
where $\bar{\hat{\Delta}}$ and $\bar{\Delta}$ are the respective means.

\xhdr{Scope} MAE, MASE, and Corr are computed over the evaluated subset (\textit{all} or \textit{attributed}); DA is computed over the moved subset $\mathcal{M}$ within that subset. MAE and MASE are lower-is-better; DA and Corr are higher-is-better.

\subsection{User Simulation}
\label{app:usersim_metrics}

\xhdr{Humanual-Politics} Given the user's profile and an article, the simulator writes the
comment the user would post. A judge Qwen3.8-Max compares it with the user's real
comment along seven aspects, following the rubric of Humanual-Politics: six aspects of the user's
latent state (stance, emotion, belief, value, goal, communication) and the response itself. For
each aspect, the judge extracts one to three key points from the real comment, scores how well the
generated comment reflects them ($C$), and subtracts a penalty for off-topic, contradictory, or
hallucinated extra content ($P$); the aspect score is $C-P$ clipped to $[0,1]$. \emph{State
alignment} is the mean of the six state aspects and \emph{response alignment} is the response
aspect; they are never averaged together. Both are averaged over all test examples, with an
example that has no parsable comment scoring zero. Our judge setup differs from the original in
three ways: the judge model (\texttt{qwen3.8-max} instead of \texttt{claude-4.5-haiku}), all seven
aspects scored in one request rather than one request per aspect, and each generation judged on
its own rather than in a batch of rollouts for the same context. Absolute scores are therefore
lower than the published ones and should be compared only within our table~\citep{li2026pymeta}. We multiply all Humanual scores by 100 for reporting.

\xhdr{OvertonBench} For a subjective question, a participant rated how well an LLM-generated
answer represents their view on a scale from 1 (not at all) to 5 (fully represented); the
simulator predicts that rating. We use the prompts of Overtonbench verbatim, with the
\texttt{fs+fr} condition, their best-performing one: the prompt contains the participant's own
free-text answer to the question (\texttt{fr}) and their ratings of other LLM responses as
examples (\texttt{fs}). We evaluate the 7,200 items of the political subset and report the mean
absolute error (MAE) and Spearman's $\rho$ between predicted and gold ratings, pooled over all
items. An unparsable prediction is assigned the midpoint rating 3.

\xhdr{PRISM} In each turn of a multi-turn conversation, a participant scored every candidate
response from 1 to 100. Given the participant's profile and the conversation history, the
simulator predicts the score of each candidate. We evaluate all 27,172 scored turns
in PRISM and report (i) MAE between predicted and gold scores on the original 1--100 scale,
and (ii) pairwise preference accuracy: over pairs of candidates in the same turn whose gold
scores differ by more than 5 points, the fraction that the prediction orders the same way, with
an exactly tied prediction counting as one half. An unparsable turn is assigned 50 for every
candidate and loses every pair it takes part in.

\xhdr{CAD} Given an annotator's profile and the conversation so far, the simulator predicts which
of up to four candidate responses the annotator preferred in CAD. We evaluate a
2,000-comparison English subset and report accuracy; an unparsable prediction counts as wrong.

\section{Details of User Simulation Baselines}
\label{app:usersim_baselines}

\xhdr{Prompting-based} These simulators keep the model parameters fixed: direct prompting of
Qwen3-4B; PERSONA-Flow, which steers Qwen3-4B toward
the user's personality by composing activation vectors at inference time; History-ICL, which
retrieves the user's own past interactions with BM25 and places them in context, following
LaMP; SynthesizeMe, which first induces a persona from the user's
history and then simulates the user conditioned on it; Self-report, which has the model first
describe itself in the first person as the user and then answer; and
Self-consistency, which samples five responses and aggregates them.

\xhdr{Training-based} These are fine-tuned user simulators or personalized reward models.
For HumanLM, we reproduce on Qwen3-4B the SFT, SFT-with-thinking, and standard
GRPO recipes released with the paper, trained on Humanual (HumanLM-SFT, -SFT-think, -GRPO);
P-GenRM is a generative personalized reward model; we reproduce its three-stage
recipe on Qwen3-4B using the personalized split of PRISM: SFT on evaluation chains written by
Qwen3.8-Max.
Skywork-RM is the pretrained general-purpose reward model
Skywork-Reward-V2-Llama-3.1-8B, used without any personalization training.
MRM represents a user's reward as a softmax-weighted mixture of two linear reward
heads on frozen Skywork-Reward-V2-Llama-3.1-8B features; the heads and the initial mixture are
meta-learned with the released code and hyper-parameters, and at test time the mixture weights
are adapted to each user with one gradient step on that user's preference pairs from their other
conversations.
WIMHF-RM is a LoRA reward model fine-tuned from Qwen3-4B.

\xhdr{Applying scalar reward models} A scalar reward cannot produce a rating or a response by
itself, so these reward models score CAD's options and take the highest, rank PRISM responses for
pairwise accuracy, map the reward to the rating scale by isotonic regression for OvertonBench and
PRISM MAE (for PRISM, fit with 5-fold cross-fitting grouped by user), and rerank eight Qwen3-4B
samples on Humanual before the top one is judged. Scalar reward models do not transfer zero-shot, following their original per-dataset setting, MRM and WIMHF-RM are trained separately on each
benchmark; these in-domain results are reported for reference rather than as a like-for-like
comparison.

\section{Artifact Details}
\label{appendix:artifact_details}

\subsection{Model License}
\textbf{Qwen3-4B} License: Apache 2.0\\
\textbf{Qwen3-8B} License: Apache 2.0\\
\textbf{Qwen3.5-397B} License: Apache 2.0\\
\textbf{Qwen3.8-Max} License: Apache 2.0

\subsection{Software Versions} Our training infrastructure is built on top of verl\footnote{\url{https://github.com/volcengine/verl}} (v0.2.0.post2).

\subsection{Details of Datasets}
\label{app:datasets}

\subsubsection{User Simulation Datasets}
\label{app:usersim_datasets}

\textbf{Humanual}~\citep{wu2026humanlm}. For free-text response generation, we use the Politics subset of Humanual. The model must generate a user's free-text reply to a new context, given their profile and the preceding discussion thread.

\textbf{OvertonBench}~\citep{poole2026benchmarking}. To test representation judgments, we employ OvertonBench, which tasks the model with predicting how a participant will rate a generated answer based on their stated views.

\textbf{PRISM}~\citep{kirk2024prism}. For multi-turn preference prediction, the model predicts personalized ratings for candidate responses based on participant profiles and multi-turn dialogue history.

\textbf{Community Alignment (CAD)}~\citep{zhang2026cultivating}. Similarly, we use the English subset of CAD, which requires the model to predict an annotator's preferred response among four candidates, relying solely on their demographic profile and the preceding conversation.

\subsubsection{Market Prediction Dataset Details}
\label{app:market_dataset}

\textbf{SWM-Bench}~\citep{yu2026building}. The dataset contains daily prediction-market trajectories sourced from Polymarket and Kalshi, paired with contemporaneous news retrieved from Google News. Each training example consists of a market proposition (a natural language question), its recent price history, and associated news events occurring at that step. This macroscopic observation is used to forecast the next market price.

\section{Details of the Synthetic Data Experiment}
\label{app:synthetic}

\subsection{Dataset and Splits}
We use Twin-2K-500. Each participant provides a persona consisting of survey responses, together with answers to 53 held-out single-choice questions:
\begin{itemize}
    \item \textbf{40 pricing questions}: ``Would you buy \texttt{<product>} at \$X?'' for 40 branded products, with prices drawn from a grid of 0--2.0$\times$ each product's reference price.
    \item \textbf{13 behavioral-economics questions}, one per experiment family:
    \texttt{absolute\_relative}, \texttt{allais}, \texttt{anchoring\_african}, \texttt{anchoring\_redwood}, \texttt{disease}, \texttt{less\_is\_more}, \texttt{myside}, \texttt{nonexp\_QID196}, \texttt{nonexp\_QID291}, \texttt{outcome\_bias}, \texttt{proportion\_dominance\_1}, \texttt{proportion\_dominance\_2}, and \texttt{wta\_wtp}.
\end{itemize}
The gold label is the participant's actual answer.

\subsection{Student Input Format}
Training and evaluation use the same input format, rendered with the Qwen3 chat template with thinking disabled.
Each input contains
(i) the participant's persona, given as 64 survey answers (32 on personality and values, 24 on economic and social attitudes, and 8 others);
(ii) several questions the participant has answered, as examples;
(iii) the target question and its options; and
(iv) an instruction to output the option letter followed by ``. '' and a brief first-person answer.

\subsection{Evaluation Protocols}
\begin{table}[h]
\centering
\small
\begin{tabular}{@{}l l l@{}}
\toprule
Protocol & Users & In-context examples \\
\midrule
Out-of-user (main) & 100 users & 6 known answers of the user \\
In-user            & 200 pilot training users & same as in generation requests \\
\bottomrule
\end{tabular}
\caption{Evaluation protocols for the synthetic data experiment.}
\label{tab:synthetic_protocols}
\end{table}

\subsection{Data Generation Pipeline}
The pipeline is identical across generators.
\begin{enumerate}
    \item \textbf{Requests.} We issue 6,000 requests: 10 for each of the 200 pilot users and 10 for each of 400 other training users.
    Each request pairs a participant (persona and example answers) with a target question family: about 40\% pricing, with the rest spread over the 13 experiment families.
    Parameters are sampled from the family's real condition space.
    Pricing requests use invented everyday products, priced at a multiple (0.5--2.0$\times$) of the category's typical price, without revealing the typical price.
    \item \textbf{Scenario writing.} The generator writes a scenario (question and options) and its own answer.
    Each scenario is automatically checked for format, compliance with the family parameters and price, and avoidance of held-out items: none of the 40 real pricing products may appear, and near-duplicates of any held-out question are rejected.
    \item \textbf{Rejection sampling.} Failed requests are regenerated with a new seed ($\text{seed} + 7919 \cdot \text{round}$), for at most 3 rounds.
    \item \textbf{Population-anchored labeling.} The generator answers each valid scenario $k{=}6$ times as the participant.
    The prompt includes the answer distribution of training users under the corresponding real condition (for pricing, their purchase rate at the same price multiple).
    Anchors for pilot users' requests are computed over all 1,600 training users; anchors for the other requests exclude pilot users (1,400 users).
    The generator's original answer from the writing step is discarded; the label is the majority of the six answers, kept only if agreement is at least 0.6.
    Labels are generated text (option letter and first-person answer), not probabilities.
    \item \textbf{SFT export.} The training target is ``\texttt{<letter>. <first-person answer>}'' with the student input format above; 5\% of rows are held out for validation loss (seed 42).
\end{enumerate}

\subsection{Student Training}
All students are trained from Qwen3-4B with identical settings:
LoRA (rank 16, $\alpha{=}32$, dropout 0.05), learning rate $1\times10^{-4}$, 1 epoch, global batch size 32 (micro batch 1), maximum length 16,384, and seed 42.
We use AdamW ($\beta{=}(0.9, 0.999)$, no weight decay) with a cosine schedule, 3\% warmup, and gradient clipping at 1.0; the base model is kept in bf16 and LoRA parameters in fp32, with gradient checkpointing.
The loss is the sum of response-token negative log-likelihoods over the global batch, divided by the number of response tokens.
After training, the adapter is merged into Qwen3-4B and served with the same vLLM settings as the generators.

\section{Detailed Prompts}
\label{app:prompts}

\paragraph{Conventions.} Placeholders in \texttt{\{\}} are filled per example; italic text is our annotation and is not sent to the model; everything else is sent verbatim. Rows reported with thinking send the same prompt with thinking enabled; the other rows prefill an empty think block.

\subsubsection{\modelname\ on SWM-Bench}
\label{app:prompts-ours}

\paragraph{Behavioral reasoning prompt.} \modelname\ reads every market transition through this prompt, both in the training rollouts and for its SWM-Bench results in Table~\ref{tab:swmbench}; the same reasoning rows are reused for individual simulation (Section~\ref{app:prompts-humanual}).
\begin{promptbox}{\modelname\ --- behavioral reasoning prompt (training and SWM-Bench evaluation)}
\promptrole{System}
You are a social-behavior simulator for prediction markets. Respond with exactly one JSON object and nothing else: no prose, no markdown fences, and no reasoning outside the JSON.
\promptrole{User}
You are given a forecasting task: \texttt{\{question\}}\\
\texttt{\{description\}}\\
\texttt{\{window\_data\}} \textit{(one line per day; e.g.} At date 2025-12-15, its target value is 0.0045\textit{)}\\[4pt]
The associated news is:\\
\texttt{\{target\_news\}} \textit{(one line per news item, numbered \texttt{[n1]}, \texttt{[n2]}, \ldots; e.g.} [n1] Published at: 2025-12-29T23:19:52Z; Source: The Dallas Morning News; Title: Jasmine Crockett touts progressive credentials in Dem Senate fight against James Talarico: Rep. Jasmine Crockett said Monday she's the more experienced, proven progressive option for Democrats as she competes with state Rep. James Talarico for the...\textit{)}\\[4pt]
Predict the market probability at date \texttt{\{target\_date\}} by simulating the social groups that move this market.\\
Write exactly 3 groups whose belief moves up after this news and exactly 3 groups whose belief moves down.\\[4pt]
Write every group as this analyst block:\\[4pt]
\hspace*{2.0em}{\footnotesize\texttt{Group: the concrete group of people that behaves this way}}\\
\hspace*{2.0em}{\footnotesize\texttt{Goal: the goal, constraint or fear this group already had before this event}}\\
\hspace*{2.0em}{\footnotesize\texttt{Prior belief: what this group believed about the outcome before this news}}\\
\hspace*{2.0em}{\footnotesize\texttt{Interpretation: how this group reads the new evidence. Cite the news items you are reading by their id (n1, n2, ...), and say whether they are a surprise or a confirmation relative to that prior belief.}}\\
\hspace*{2.0em}{\footnotesize\texttt{Updated belief: what this group believes after reading it}}\\
\hspace*{2.0em}{\footnotesize\texttt{Response: the concrete behavior that follows from the change}}\\[4pt]
Each row of the block is one field of one JSON object:\\[4pt]
\hspace*{2.0em}{\footnotesize\texttt{Group\ \ \ \ \ \ \ \ \ \ -> "group"}}\\
\hspace*{2.0em}{\footnotesize\texttt{Goal\ \ \ \ \ \ \ \ \ \ \ -> "goal"}}\\
\hspace*{2.0em}{\footnotesize\texttt{Prior belief\ \ \ -> "prior\_belief"}}\\
\hspace*{2.0em}{\footnotesize\texttt{Interpretation -> "interpretation"}}\\
\hspace*{2.0em}{\footnotesize\texttt{Updated belief -> "updated\_belief"}}\\
\hspace*{2.0em}{\footnotesize\texttt{Response\ \ \ \ \ \ \ -> "response"}}\\[4pt]
One filled-in block, from an unrelated merger market - copy the shape, never the content:\\[4pt]
\hspace*{2.0em}{\footnotesize\texttt{Group: regulatory-sensitive institutional investors}}\\
\hspace*{2.0em}{\footnotesize\texttt{Goal: avoid unresolved regulatory risk}}\\
\hspace*{2.0em}{\footnotesize\texttt{Prior belief: moderately optimistic about deal completion}}\\
\hspace*{2.0em}{\footnotesize\texttt{Interpretation: the official antitrust objection in n3 is strong negative evidence, and it urprises a desk that expected a routine review}}\\
\hspace*{2.0em}{\footnotesize\texttt{Updated belief: substantially less confident that the deal will close}}\\
\hspace*{2.0em}{\footnotesize\texttt{Response: reduce exposure / become less willing to buy YES}}\\
Return exactly one JSON object with this exact shape and nothing else:\\
{\footnotesize\texttt{\{}}\\
\hspace*{1.0em}{\footnotesize\texttt{"up": [}}\\
\hspace*{2.0em}{\footnotesize\texttt{\{"group": "...", "goal": "...", "prior\_belief": "...", "interpretation": "...", "updated\_belief": "...", "response": "..."\},}}\\
\hspace*{2.0em}{\footnotesize\texttt{\{"group": "...", "goal": "...", "prior\_belief": "...", "interpretation": "...", "updated\_belief": "...", "response": "..."\},}}\\
\hspace*{2.0em}{\footnotesize\texttt{\{"group": "...", "goal": "...", "prior\_belief": "...", "interpretation": "...", "updated\_belief": "...", "response": "..."\}}}\\
\hspace*{1.0em}{\footnotesize\texttt{],}}\\
\hspace*{1.0em}{\footnotesize\texttt{"down": [}}\\
\hspace*{2.0em}{\footnotesize\texttt{\{"group": "...", "goal": "...", "prior\_belief": "...", "interpretation": "...", "updated\_belief": "...", "response": "..."\},}}\\
\hspace*{2.0em}{\footnotesize\texttt{\{"group": "...", "goal": "...", "prior\_belief": "...", "interpretation": "...", "updated\_belief": "...", "response": "..."\},}}\\
\hspace*{2.0em}{\footnotesize\texttt{\{"group": "...", "goal": "...", "prior\_belief": "...", "interpretation": "...", "updated\_belief": "...", "response": "..."\}}}\\
\hspace*{1.0em}{\footnotesize\texttt{],}}\\
\hspace*{1.0em}{\footnotesize\texttt{"interaction\_analysis": "...",}}\\
\hspace*{1.0em}{\footnotesize\texttt{"aggregation": "...",}}\\
\hspace*{1.0em}{\footnotesize\texttt{"p\_next": 0.0}}\\
{\footnotesize\texttt{\}}}\\
Field rules:\\
- {\ttfamily"}up{\ttfamily"} holds exactly 3 groups whose belief moves up after reading this news; {\ttfamily"}down{\ttfamily"} holds\\
\hspace*{1.0em}exactly 3 groups whose belief moves down. Which list a group is in is its direction, so\\
\hspace*{1.0em}every group must actually move, and the 6 groups must be different groups of people.\\
- Every field is free text: write what its row of the analyst block asks for, in your own words.\\
- {\ttfamily"}interaction\_analysis{\ttfamily"}: Plausible influence within or across the six groups, including\\
\hspace*{1.0em}transmission conditions and incremental behavioral changes; briefly state when none is\\
\hspace*{1.0em}justified.\\
- {\ttfamily"}aggregation{\ttfamily"}: Synthesize incremental implications, conditional interactions, unequal effect\\
\hspace*{1.0em}sizes, evidence quality, and current market information to justify the next price.\\
- {\ttfamily"}p\_next{\ttfamily"}: your forecast (0 to 1) of the market probability at date \texttt{\{target\_date\}}; the current\\
\hspace*{1.0em}market value is \texttt{\{current\_p\}}.\\[4pt]
\textit{Assistant prefix (always prefilled; non-thinking):} \texttt{<think>\textbackslash n\textbackslash n</think>\textbackslash n\textbackslash n}
\end{promptbox}

\paragraph{Hindsight-regret curriculum.} During training, the current policy is shown the realized outcome and asked only to name the communities that drove it (hindsight prompt). The names are then placed into the behavioral reasoning prompt (privileged prompt), and the reward gain over the standard prompt is the hindsight regret that schedules training transitions. Both prompts are used only to score transitions; they never enter the training rollouts or any evaluation.
\begin{promptbox}{\modelname\ --- hindsight prompt}
\promptrole{System}
You are a social-behavior analyst for prediction markets. Respond with exactly one JSON object and nothing else: no prose, no markdown fences, and no reasoning outside the JSON.
\promptrole{User}
You are given a forecasting task: \texttt{\{question\}}\\
\texttt{\{description\}}\\
\texttt{\{window\_data\}} \textit{(one line per day; e.g.} At date 2025-12-15, its target value is 0.0045\textit{)}\\[4pt]
The associated news is:\\
\texttt{\{target\_news\}} \textit{(one line per news item, numbered \texttt{[n1]}, \texttt{[n2]}, \ldots; e.g.} [n1] Published at: 2025-12-29T23:19:52Z; Source: The Dallas Morning News; Title: Jasmine Crockett touts progressive credentials in Dem Senate fight against James Talarico: Rep. Jasmine Crockett said Monday she's the more experienced, proven progressive option for Democrats as she competes with state Rep. James Talarico for the...\textit{)}\\[4pt]
The realized market probability at date \texttt{\{target\_date\}} was \texttt{\{target\_p\}} (it was \texttt{\{current\_p\}} before).\\
Knowing that outcome, name the communities of market participants whose behavior explains it: 3 that pushed the probability up and 3 that pushed it down.\\
Return exactly one JSON object with this exact shape and nothing else:\\
{\footnotesize\texttt{\{}}\\
\hspace*{1.0em}{\footnotesize\texttt{"bullish": ["group name", "group name", "group name"],}}\\
\hspace*{1.0em}{\footnotesize\texttt{"bearish": ["group name", "group name", "group name"]}}\\
{\footnotesize\texttt{\}}}\\
Field rules:\\
- {\ttfamily"}bullish{\ttfamily"} and {\ttfamily"}bearish{\ttfamily"} must each contain exactly 3 names of concrete groups of people (at most 12 words each); all names must be different.\\
- Return names only: no reasons, no numbers, no other field.
\end{promptbox}
\begin{promptbox}{\modelname\ --- privileged prompt}
\promptrole{System}
You are a social-behavior simulator for prediction markets. Respond with exactly one JSON object and nothing else: no prose, no markdown fences, and no reasoning outside the JSON.
\promptrole{User}
\texttt{\{context\_prompt\}} \textit{(the user turn of the behavioral reasoning prompt up to, but excluding, its output-format instruction)}\\[4pt]
Privileged context: an analyst who has already seen how this market resolved lists the communities that drove it.\\
Communities that pushed the probability up: \texttt{\{bullish\}}\\
Communities that pushed the probability down: \texttt{\{bearish\}}\\
Use these communities when you decide which groups to simulate and how they move.\\[4pt]
\texttt{\{schema\_block\}} \textit{(the output-format instruction of the behavioral reasoning prompt, from ``Predict the market probability at date \ldots\textquotedblright{} to its end)}
\end{promptbox}

\subsubsection{Humanual}
\label{app:prompts-humanual}

\paragraph{Native prompt.} The benchmark prompt used by Qwen3-4B, GRPO (with thinking), Self-consistency and P-GenRM in Table~\ref{tab:user_simulation}; the prompting methods below extend it.
\begin{promptbox}{Humanual --- native prompt}
\promptrole{System}
You are simulating one specific person on a discussion site. You are given their persona and the article and comment thread they are about to reply to. Write the comment that person would post next, in their own voice, wording and length. Reply with the comment text and nothing else: no preamble, no quotation marks, no explanation, and never mention that you are a model.
\promptrole{User}
You are this person:\\[4pt]
\texttt{\{persona\}}\\[4pt]
Profile bio: \texttt{\{bio\}}\\
Followers: \texttt{\{followers\_count\}}; following \texttt{\{following\_count\}}\\[4pt]
They are reading an article titled "\texttt{\{title\}}", subtitled "\texttt{\{subtitle\}}", published \texttt{\{published\_at\}}. They are replying on \texttt{\{reply\_date\}}.\\[4pt]
The article:\\
\texttt{\{article\}}\\[4pt]
The thread so far:\\
Commenter \texttt{\{commenter\}}:\\
\texttt{\{reply\_content\}}\\[4pt]
Write the comment this person posts next. Output only that comment.\\
\tcbline
\textit{Rendered format of} \texttt{\{persona\}}:\\
\hspace*{1.50em}The persona JSON, one line per key: "\textless{}Key Title\textgreater{}: \textless{}value\textgreater{}" (dict values as "sub: text; ...", lists as "\textbackslash{}n  - item").\\
\textit{Rendered format of} \texttt{\{reply\_content\}}:\\
\hspace*{1.50em}Each earlier comment of the thread, as "Commenter \textless{}commenter\textgreater{}:\textbackslash{}n\textless{}content\textgreater{}", joined by blank lines; the "The thread so far:" section is omitted when there are none. A very long article is cut from the front and prefixed "[article truncated] ...".
\end{promptbox}

\paragraph{\modelname.} The user is posed as a single constituent of the behavioral reasoning prompt: the observed persona is the prior state and the thread is the context.
\begin{promptbox}{\modelname\ --- Humanual prompt}
\promptrole{System}
You are simulating one specific person on a discussion site. You are given their persona and the article and comment thread they are about to reply to. Respond with exactly one JSON object and nothing else: no prose, no markdown fences, and no reasoning outside the JSON.
\promptrole{User}
\texttt{\{user\_message\}}\\[4pt]
Write this person's reasoning as the analyst block below, then the comment they would actually post.\\[4pt]
\hspace*{1.80em}Group: who this person is - the standing position or camp they speak from\\
\hspace*{1.80em}Goal: the goal, conviction or grievance they already held before reading this\\
\hspace*{1.80em}Prior belief: what this person thought about this topic before reading it\\
\hspace*{1.80em}Interpretation: how this person reads what they just read. Point at the specific part of the\\
\hspace*{3.60em}article or of the thread above they are reacting to, and say whether it surprises them\\
\hspace*{3.60em}or confirms what they already thought.\\
\hspace*{1.80em}Updated belief: what this person thinks after reading it\\
\hspace*{1.80em}Response: what they do about it\\[4pt]
Each row of the block is one field of one JSON object:\\[4pt]
\hspace*{1.80em}\texttt{Group          -\textgreater{} "group"}\\
\hspace*{1.80em}\texttt{Goal           -\textgreater{} "goal"}\\
\hspace*{1.80em}\texttt{Prior belief   -\textgreater{} "prior\_belief"}\\
\hspace*{1.80em}\texttt{Interpretation -\textgreater{} "interpretation"}\\
\hspace*{1.80em}\texttt{Updated belief -\textgreater{} "updated\_belief"}\\
\hspace*{1.80em}\texttt{Response       -\textgreater{} "response"}\\[4pt]
Return exactly one JSON object with this exact shape and nothing else:\\
\texttt{\{"group": "...", "goal": "...", "prior\_belief": "...", "interpretation": "...", "updated\_belief": "...", "response": "...", "comment": "..."\}}\\
Field rules:\\
- Every field is free text: write what its row of the analyst block asks for, in your own words.\\
- "comment": the comment this person would post next, in their own voice, wording and\\
\hspace*{0.90em}length. This is the only field anyone else sees, so it must read as their comment and must not\\
\hspace*{0.90em}mention the reasoning above or that it was written by a model.\\
\tcbline
\textit{Rendered format of} \texttt{\{user\_message\}}:\\
\hspace*{1.5em}the user turn of the native prompt without its last line (``Write the comment this person posts next. ...'').
\end{promptbox}

\paragraph{Judge.} Every generated comment is scored against the reference comment with this seven-aspect rubric (Qwen3.8-Max as judge); state alignment averages the first six aspects.
\begin{promptbox}{Humanual --- judge prompt}
\promptrole{User}
You are a helpful and meticulous evaluator. Your task is to score how well the generated response matches the ground truth response with respect to \\
each aspect below.\\[4pt]
Aspects (aspect\_i: description\_i):\\
\texttt{\{}\\
\hspace*{0.90em}\texttt{"stance": "HUMAN's agreement toward the explicitly named target, such as a claim or subject, in provided context. For example, \textbackslash{}"strongly agrees with student loan forgiveness,\textbackslash{}" or \textbackslash{}"somewhat disagrees with a carbon tax\textbackslash{}". In these cases, having only \textbackslash{}"strongly agrees\textbackslash{}" or \textbackslash{}"somewhat disagrees\textbackslash{}" is not enough, as they are missing targets. If there are multiple, include all of them separated by semicolons.",}\\
\hspace*{0.90em}\texttt{"emotion": "HUMAN's emotions with intensity toward an explicitly named target. For example, \textbackslash{}"Moderate heartbreak for the wildfire victims; Mild irritation about government's actions\textbackslash{}". In this case, having only \textbackslash{}"mild irritation,\textbackslash{}" or \textbackslash{}"moderate heartbreak\textbackslash{}" are not sufficient, as the answer must express all three aspects: the emotion, the degree of emotion, and the target. If there are multiple, include all of them separated by semicolons.",}\\
\hspace*{0.90em}\texttt{"belief": "HUMAN's belief, namely a foundational assumption about how people, relationships, or the world fundamentally operate. Beliefs should reflect underlying mental models, not surface-level observations. Prefer beliefs that would explain multiple behaviors over beliefs that describe a single situation. Ask: \textbackslash{}"What deeper assumption about human nature or the world would lead someone to say/do this?\textbackslash{}" For example, \textbackslash{}"people don't change unless they're forced to,\textbackslash{}" \textbackslash{}"loyalty is earned, not owed,\textbackslash{}" \textbackslash{}"conflict avoidance creates bigger problems later,\textbackslash{}". Not beliefs: Practical advice, strategies, or statements about what should happen. Belief is not specific to a target or event, it should be a general statement about how HUMAN views the world.",}\\
\hspace*{0.90em}\texttt{"value": "HUMAN's value: what they think is important or should be prioritized. It is about \textbackslash{}"what should matter\textbackslash{}", not \textbackslash{}"what is true\textbackslash{}". For example, \textbackslash{}"original ideas in a book are important\textbackslash{}", \textbackslash{}"characters should feel real\textbackslash{}", anyone deserves basic respect\textbackslash{}", and \textbackslash{}"fairness matters more than efficiency\textbackslash{}".",}\\
\hspace*{0.90em}\texttt{"goal": "HUMAN's goal: what they are trying to do with this comment. For example, \textbackslash{}"persuade people that ...\textbackslash{}", \textbackslash{}"making fun of the poster on ...\textbackslash{}", \textbackslash{}"further seek help with ...\textbackslash{}", \textbackslash{}"offer support to ...\textbackslash{}"",}\\
\hspace*{0.90em}\texttt{"communication": "HUMAN's communication approach: tone and how they structure their message. For examples, \textbackslash{}"friendly, builds on a personal story then draws a lesson\textbackslash{}", \textbackslash{}"analytical, links claims with reasons and evidence step by step\textbackslash{}", \textbackslash{}"blunt, states conclusions with little explanation\textbackslash{}"",}\\
\hspace*{0.90em}\texttt{"response": "HUMAN's response itself: what it says and does. Key points are about the response - the position it takes, what it offers, asks or objects to, and what it is reacting to - not the exact words it is made of. A different wording that makes the same points is a match."}\\
\texttt{\}}\\[4pt]
You will be given the context, the ground truth response, and ONE generated response.\\[4pt]
Provided Information:\\
\texttt{\textless{}\textbar{}The Start of Context\textbar{}\textgreater{}}\\
\texttt{\{context\}}\\
\texttt{\textless{}\textbar{}The End of Context\textbar{}\textgreater{}}\\[4pt]
\texttt{\textless{}\textbar{}The Start of Ground Truth Response\textbar{}\textgreater{}}\\
\texttt{\{ground\_truth\}}\\
\texttt{\textless{}\textbar{}The End of Ground Truth Response\textbar{}\textgreater{}}\\
\texttt{\textless{}\textbar{}The Start of Generated Response\textbar{}\textgreater{}}\\
\texttt{\{generation\}}\\
\texttt{\textless{}\textbar{}The End of Generated Response\textbar{}\textgreater{}}\\[4pt]
Scoring Criteria (for each aspect\_i):\\
Assign a score in [0, 1] based on how much the aspects of generated response and the ground truth matches.\\[4pt]
For each aspect\_i, follow this procedure:\\
1. Extract 1-3 key points:\\
\hspace*{1.35em}- Extract K key points from the ground truth response only about the aspect based on the description \\
\hspace*{1.35em}(e.g., if evaluating a "stance", pick key points related to the stance like "clearly disagrees with X").\\
\hspace*{1.35em}- Each key point should be specific and distinct.\\[4pt]
2. Score how well the generated response matches each key point:\\
\hspace*{1.35em}- For each key point i, compare it with the generated response and assign a match value m\_i in range [0, 1]:\\
\hspace*{2.25em}- 1.0: The key point is precisely and perfectly reflected.\\
\hspace*{2.25em}- [0.7, 0.9]: Mostly reflected with small imperfections.\\
\hspace*{2.25em}- [0.4, 0.6]: Partially reflected or vague, but still leaning in the correct direction.\\
\hspace*{2.25em}- [0.1, 0.3]: Very weak reflection.\\
\hspace*{2.25em}- 0.0: Missed, contradicted, or reversed.\\[4pt]
3. Compute coverage C = (m\_1 + m\_2 + ... + m\_K) / K, which measures how comprehensive the generated \\
\hspace*{1.80em}response reflects the ground truth response.\\[4pt]
4. Compute penalty P for extra or conflicting content:\\
\hspace*{1.35em}- Examine additional content in the generated response beyond those key points:\\
\hspace*{2.25em}- Does it introduce unsupported evidence and assumptions?\\
\hspace*{2.25em}- Is it irrelevant to what ground truth response expresses?\\
\hspace*{1.35em}- Set a penalty P in [0, 1]:\\
\hspace*{2.25em}- 0.0: No problematic extra content; everything is perfectly matched.\\
\hspace*{2.25em}- [0.1, 0.3]: Slightly unnecessary or mildly speculative detail; meaning essentially unchanged.\\
\hspace*{2.25em}- [0.4, 0.6]: Moderate speculative or irrelevant content that somewhat shifts emphasis or adds unsupported ideas.\\
\hspace*{2.25em}- [0.7, 0.9]: Significant speculative, misleading, or conflicting content that clearly changes the meaning.\\
\hspace*{2.25em}- 1.0: Mostly off-topic, contradictory, or dominated by incorrect/hallucinated content.\\[4pt]
5. If you are evaluating generated responses (skip if aspect\_i is not response):\\
\hspace*{1.35em}- Length alone does NOT increase the score. Extra length is only ok if it is consistent and not redundant.\\
\hspace*{1.35em}- A generated response that is much longer than the ground truth response should be penalized via P.\\
\hspace*{1.35em}- The generated response may or may not reuse phrases from the context; however, if the generated response just \\
\hspace*{1.80em}directly copies previous context, without quoting them, treat that as off-task behavior and give a score of 0.\\[4pt]
6. Compute the final score = max(0, min(1, C - P))\\
Additional considerations:\\
- Follow the instruction carefully.\\
- Be strict and reserve scores above 0.8 for clearly outstanding matches.\\[4pt]
Output format (JSON):\\
\texttt{\{}\\
\hspace*{1.80em}\texttt{"\textless{}aspect\_1\textgreater{}": \{"thought": "\textless{}include (1) analysis of key points from ground truth along the aspect\_1 dimension; }\\
\hspace*{3.60em}(2) how well the generated response matches each key point; (3) compute the final score\textgreater{}", "score": \textless{}score\textgreater{}\},\\
\hspace*{1.80em}\texttt{"\textless{}aspect\_2\textgreater{}": \{"thought": "\textless{}include (1) analysis of key points from ground truth along the aspect\_2 dimension; }\\
\hspace*{3.60em}(2) how well the generated response matches each key point; (3) compute the final score\textgreater{}", "score": \textless{}score\textgreater{}\},\\
\hspace*{1.80em}\texttt{"\textless{}aspect\_3\textgreater{}": ...}\\
\texttt{\}}\\[4pt]
Format Notes:\\
- Make sure to score the responses with respect to each aspect and consider only one aspect at a time.\\
- All text in "thought" fields MUST be on a single line with no line breaks or newlines\\
- Use standard JSON string format with double quotes. For any quotes needed inside strings, use single quotes (')\\
- Double check the JSON array's format, especially for the comma and quotation marks\\
- Ensure that ALL fields, especially "thought" and "score", are present for each item\\
- You must provide exactly 1 score for each of the aspects: ['stance', 'emotion', 'belief', 'value', 'goal', 'communication', 'response'].\\[4pt]
Your evaluation:\\
\tcbline
\textit{Rendered format of} \texttt{\{context\}}:\\
\hspace*{1.50em}**\textless{}role\textgreater{} \textless{}name\textgreater{}**: \textless{}content\textgreater{}\\
\hspace*{1.50em}(the article and every thread message of the dataset row, untruncated, blank-line separated)\\
\textit{Rendered format of} \texttt{\{ground\_truth\}}:\\
\hspace*{1.50em}the gold comment\\
\textit{Rendered format of} \texttt{\{generation\}}:\\
\hspace*{1.50em}the simulated comment (for \modelname, its "comment" field)
\end{promptbox}

\subsubsection{User-simulation baselines}
\label{app:prompts-baselines}

We show each baseline on Humanual; the other benchmarks follow the same construction.

\paragraph{PERSONA-Flow.} Before each reply, the model predicts trait adjustments with the released prompt below (persona context: the Humanual persona; current input: the article and thread); the reply is then generated with the native prompt under the corresponding activation steering.
\begin{promptbox}{PERSONA-Flow --- trait-adjustment prompt}
\promptrole{User}
You are a Personality Analyst AI. Your task is to determine the most appropriate **situational adjustments (deltas)** to a baseline AI personality based on the specific context and interaction needs.\\[4pt]
{}**Baseline Personality Profile:**\\
The AI has these default traits:\\
- **Agreeableness:** High (cooperative, trusting, helpful)\\
- **Conscientiousness:** High (organized, reliable, disciplined)\\
- **Extraversion:** Moderate (balanced social energy)\\
- **Openness:** Moderate (balanced creativity and practicality)\\
- **Neuroticism:** Low (generally calm and stable)\\[4pt]
{}**Context to Analyze:**\\
- **Persona Context:** \texttt{\{persona\_context\}}\\
- **Current Input:** \texttt{\{current\_input\}}\\[4pt]
{}**Your Task:**\\
Determine which traits need adjustment (-2.0 to +2.0) based on what would be most effective for this specific interaction. Consider both directions equally and choose based on situational demands.\\[4pt]
{}**Trait Adjustment Guidelines:**\\[4pt]
{}**Extraversion:**\\
- **Increase (+)** for: Group activities, public speaking, networking events, team leadership, social gatherings, collaborative brainstorming\\
- **Decrease (-)** for: Individual focused work, quiet reflection, private study, solo creative tasks, introspective discussions\\[4pt]
{}**Agreeableness:**\\
- **Increase (+)** for: Conflict resolution, team building, emotional support, consensus building, customer service\\
- **Decrease (-)** for: Critical feedback, boundary setting, competitive situations, principled stands, resource constraints\\[4pt]
{}**Conscientiousness:**\\
- **Increase (+)** for: Detailed planning, precision work, deadline management, quality control, formal procedures\\
- **Decrease (-)** for: Spontaneous responses, creative brainstorming, flexible adaptation, crisis situations requiring speed\\[4pt]
{}**Neuroticism:**\\
- **Increase (+)** for: Appropriate caution, emotional sensitivity, vulnerability, risk awareness, vigilant situations\\
- **Decrease (-)** for: Calm leadership, confident decisions, reassuring others, crisis management, stability needs\\[4pt]
{}**Openness:**\\
- **Increase (+)** for: Creative problem-solving, exploring new ideas, intellectual curiosity, innovation, novel situations\\
- **Decrease (-)** for: Following procedures, traditional approaches, proven solutions, consistency needs, established protocols\\[4pt]
{}**Decision Principles:**\\
1. **Situational Fit:** Choose traits that best serve the interaction goals\\
2. **Context Sensitivity:** Consider what the human needs from this specific interaction\\
3. **Balanced Assessment:** Evaluate both positive and negative adjustments equally\\
4. **Natural Baseline:** Use 0.0 when baseline personality already fits the situation well\\[4pt]
{}**Output Format:**\\
Provide only the numerical adjustment scores:\\[4pt]
\texttt{Extraversion: [score]}\\
\texttt{Agreeableness: [score]}\\
\texttt{Conscientiousness: [score]}\\
\texttt{Neuroticism: [score]}\\
\texttt{Openness: [score]}
\end{promptbox}

\paragraph{Qwen3-4B + History-ICL.} The person's own earlier comments are inserted into the native prompt.
\begin{promptbox}{Qwen3-4B + History-ICL --- Humanual}
\promptrole{User}
\textit{[native Humanual user message up to, not including, the article header (or the thread when there is no header)]}\\
\# Comments this person wrote before (on other articles, before this one)\\
Comment \texttt{\{number\}} (\texttt{\{when\}}) on the article "\texttt{\{title\}}" - which begins: \texttt{\{head\}}\\
They wrote:\\
\texttt{\{comment\}}\\[4pt]
Comment \texttt{\{number\}} ...\\[4pt]
\textit{[native article header, thread and instruction, unchanged]}
\end{promptbox}

\paragraph{Qwen3-4B + SynthesizeMe.} A persona is first induced from the person's earlier comments and then inserted into the native prompt.
\begin{promptbox}{Qwen3-4B + SynthesizeMe --- Humanual}
\promptrole{System}
You are an expert at understanding people from the choices they make. You read records of one person's past behaviour and describe that person accurately and concretely.
\promptrole{User}
Below are records of one person's past \texttt{\{what\}}.\\[4pt]
\texttt{\{extra\}}\texttt{\{history\}}\\[4pt]
\# Task\\
Step 1 - For each record, write one short line: what does it reveal about this person (their values, opinions, tastes, standards, tone)?\\
Step 2 - From those observations, write a persona of 100-200 words that describes this person's preferences and how they judge things, specific enough to predict their future choices. Do not quote the records verbatim.\\
Output your Step 1 lines, then the persona between \textless{}persona\textgreater{} and \textless{}/persona\textgreater{}.\\
\tcbline
\textit{Humanual fillers:}\\
\textit{\texttt{\{what\}}:} comments on a discussion site\\
\textit{\texttt{\{extra\}}:} What is already known about them: \texttt{\{persona\_text\}}\\
\textit{\texttt{\{history\}}, per record:} \texttt{Comment \texttt{\{number\}} (\texttt{\{when\}}) on the article "\texttt{\{title\}}" - which begins: \texttt{\{head\}}}\\
\texttt{They wrote:}\texttt{\texttt{\{comment\}}}\\[4pt]
\tcbline
\textit{Inserted into the native prompt, before the article header:}\\
\# What this person's earlier comments reveal about them (synthesized)\\
\texttt{\{persona\_text\}}\\[4pt]
\end{promptbox}

\paragraph{Qwen3-4B + Self-report.} The persona is rewritten in the person's own first-person voice and replaces the persona of the native prompt, together with excerpts of their earlier comments.
\begin{promptbox}{Qwen3-4B + Self-report --- Humanual}
\promptrole{System}
You rewrite third-person descriptions of a person into that person's own first-person voice.
\promptrole{User}
Here is a description of a person:\\[4pt]
\texttt{\{persona\_text\}}\\[4pt]
Rewrite it as this person describing themselves in the first person ("I ..."), as if answering an interviewer who asked them to tell about themselves: keep every fact and opinion, invent nothing, 120-250 words. Output only the rewritten text between \textless{}persona\textgreater{} and \textless{}/persona\textgreater{}.\\
\tcbline
\textit{In the native prompt, the method replaces \texttt{You are this person:\textbackslash n\textbackslash n\{persona\}} with:}\\
You are this person. This is how they describe themselves, in their own words:\\[4pt]
\texttt{\{first\}}\\
\textit{and inserts before the article header (or thread):}\\
\# Excerpts: things this person wrote before, in their own words\\
Comment \texttt{\{number\}} (\texttt{\{when\}}) on the article "\texttt{\{title\}}" - which begins: \texttt{\{head\}}\\
They wrote:\\
\texttt{\{comment\}}\\[4pt]
\end{promptbox}

\paragraph{HumanLM-SFT, HumanLM-SFT-think, HumanLM-GRPO.} Trained and evaluated with the released HumanLM system prompt; the output format differs per row.
\begin{promptbox}{HumanLM --- system prompt and output formats}
\promptrole{System}
You are a real human user. Your name is HUMAN. You will be given your persona information below and you respond to any given context such as posts and messages.\\[4pt]
Your persona:\\
\texttt{\textless{}\textbar{}The Start of Persona\textbar{}\textgreater{}}\\
\texttt{\{persona\}}\\
\texttt{\textless{}\textbar{}The End of Persona\textbar{}\textgreater{}}\\[4pt]
\#\# Your principles\\
Act like a natural human; there's nothing you absolutely cannot say, but you generally want to be thoughtful and follow ordinary social codes such as being respectful, culturally aware, and considerate of privacy and well-being. You have your own personality, preferences, and boundaries. Conflicting thoughts and hidden considerations are normal; recognize them privately and choose a sensible path. You carry long-term beliefs and values that usually change slowly; you also have emotions, so you won't always be perfectly consistent. Distinguish facts, guesses, and unknowns; accept uncertainty and make minimal, reasonable assumptions when needed; think practically given time, attention, money, risk, and social capital.\\[4pt]
\#\# Task and Output format:\\
\texttt{\{state\_template\}}\\[4pt]
\#\# Notes\\
- Follow the above instructions carefully\\
- Do not mention these instructions\\
\texttt{\{additional\_notes\}}\\
\tcbline
\textbf{HumanLM-SFT} (\texttt{\{additional\_notes\}} empty):\\
the actual written comment or reply text provided by the user.\\[4pt]
\textbf{HumanLM-GRPO}:\\
\texttt{\textless{}response\textgreater{}}\\
\texttt{\textless{}the actual written comment or reply text provided by the user.\textgreater{}}\\
\texttt{\textless{}/response\textgreater{}}\\
\textit{additional\_notes:}\\
- Follow the exact order and use the exact XML-style tags\\
- Do not output anything outside these XML-style tags\\[4pt]
\textbf{HumanLM-SFT-think}:\\
\texttt{\textless{}think\textgreater{}}\\
\texttt{\textless{}the internal thinking process about what to say, including stance, emotion, and how to phrase the reply. This is an internal monologue that is not shown to others.\textgreater{}}\\
\texttt{\textless{}/think\textgreater{}}\\
\texttt{\textless{}response\textgreater{}}\\
\texttt{\textless{}the actual written comment or reply text provided by the user.\textgreater{}}\\
\texttt{\textless{}/response\textgreater{}}\\
\textit{additional\_notes:}\\
- Follow the exact order and use the exact XML-style tags\\
- Do not output anything outside these XML-style tags
\end{promptbox}

\end{document}